\documentclass[10pt,twocolumn,letterpaper]{article}

\usepackage[pagenumbers]{cvpr} 

\definecolor{darkgreen}{RGB}{0,128,0}

\usepackage{cuted}
\usepackage{multirow}
\newcommand{\MID}{M-ID}
\newcommand{\MIDD}{M-ID-$\Delta$}
\newcommand{\mMID}{\textnormal{M-ID}}

\usepackage{iftex}
\ifTUTeX
  \makeatletter
  \@ifpackageloaded{fontspec}{}{%
    \usepackage[no-math]{fontspec}%
    \setmainfont{texgyretermes}[Extension=.otf,UprightFont=*-regular,
      BoldFont=*-bold,ItalicFont=*-italic,BoldItalicFont=*-bolditalic]%
    \setsansfont{texgyreheros}[Extension=.otf,UprightFont=*-regular,
      BoldFont=*-bold,ItalicFont=*-italic,BoldItalicFont=*-bolditalic]%
    \setmonofont{texgyrecursor}[Extension=.otf,UprightFont=*-regular,
      BoldFont=*-bold,ItalicFont=*-italic,BoldItalicFont=*-bolditalic]}
  \makeatother
\fi

\usepackage{amsthm}
\providecommand{\propositionname}{Proposition}
\makeatletter
\newtheoremstyle{tightplain}{4pt}{4pt}{\itshape}{}{\bfseries}{.}{.5em}{}
\newtheoremstyle{tightdefinition}{4pt}{4pt}{\normalfont}{}{\bfseries}{.}{.5em}{}
\@ifpackageloaded{ctex}{\theoremstyle{tightdefinition}}{\theoremstyle{tightplain}}
\renewenvironment{proof}[1][\proofname]{\par
  \pushQED{\qed}%
  \normalfont \topsep4\p@ \partopsep\z@\relax
  \trivlist
  \item\relax
  {\itshape #1\@addpunct{.}}\hspace\labelsep\ignorespaces
}{\popQED\endtrivlist\@endpefalse}
\makeatother
\newtheorem{proposition}{\propositionname}
\AtEndPreamble{%
  \crefname{proposition}{Prop.}{Props.}%
  \Crefname{proposition}{Proposition}{Propositions}}

\usepackage{fontawesome5}
\usepackage{pifont}

\newcommand{\rf}[1]{\textcolor{gray}{#1}}

\usepackage{wrapfig}

\usepackage{colortbl}

\definecolor{figNavy}{HTML}{23395B}
\definecolor{figCoral}{HTML}{E4572E}
\definecolor{figMuted}{HTML}{8E8A80}

\definecolor{figMutedDark}{HTML}{5E5A52}

\definecolor{figPurple}{HTML}{7A5C99}
\definecolor{figTeal}{HTML}{2A9D8F}
\definecolor{figGold}{HTML}{D99A1C}
\definecolor{figCoralMid}{HTML}{EE8A69}
\definecolor{figCoralLight}{HTML}{F5BCA6}
\newcommand{\legLine}[1]{\textcolor{#1}{\rule[0.45ex]{1.1em}{1.1pt}}}
\newcommand{\legStart}{\textcolor{figNavy}{\large$\bullet$}}
\newcommand{\legGoal}{\textcolor{figGold}{$\bigstar$}}

\usepackage{tikz}
\usetikzlibrary{patterns}
\usetikzlibrary{arrows.meta,calc,decorations.pathreplacing,shapes.geometric} 
\definecolor{surpriseGray}{HTML}{7C8490}
\definecolor{surprisePurple}{HTML}{8073AC}
\definecolor{surpriseOrange}{HTML}{E76F51}
\DeclareRobustCommand{\legSolidDot}{\tikz[baseline=-0.6ex]{\draw[figNavy,line width=0.8pt] (0,0) -- (1.4em,0);\fill[figNavy] (0.7em,0) circle (1.2pt);}}
\DeclareRobustCommand{\legDashedOpen}{\tikz[baseline=-0.6ex]{\draw[figNavy,line width=0.8pt,dash pattern=on 2pt off 1.2pt] (0,0) -- (1.4em,0);\draw[figNavy,fill=white,line width=0.65pt] (0.7em,0) circle (1.2pt);}}
\DeclareRobustCommand{\legRadius}{\tikz[baseline=-0.6ex]\draw[figNavy,line width=0.5pt,dash pattern=on 0.6pt off 0.7pt] (0,0) circle (0.85ex);}
\newcommand{\legBox}[1]{\textcolor{#1}{\rule{0.75em}{0.75em}}}
\definecolor{figLight}{HTML}{D6D1C4}
\definecolor{figMid}{HTML}{B5AFA2}
\DeclareRobustCommand{\legHatch}[1]{\tikz[baseline=-0.1ex]{\fill[#1] (0,0) rectangle (0.75em,0.75em);\fill[pattern=north east lines,pattern color=white] (0,0) rectangle (0.75em,0.75em);}}
\definecolor{figPaper}{HTML}{D9D4C8}
\definecolor{figSandA}{HTML}{E6E1D7}
\definecolor{figSandB}{HTML}{C9C2B4}
\definecolor{figSandC}{HTML}{A69F90}
\DeclareRobustCommand{\legDots}[1]{\tikz[baseline=-0.1ex]{\fill[#1] (0,0) rectangle (0.75em,0.75em);\fill[pattern=dots,pattern color=white] (0,0) rectangle (0.75em,0.75em);}}
\definecolor{figRoute}{HTML}{1A1A1A}

\DeclareRobustCommand{\legWalkStart}{\tikz[baseline=-0.6ex]\fill[figRoute] (0,0) circle (2.1pt);}
\DeclareRobustCommand{\legWalkGoal}{\tikz[baseline=-0.6ex]\node[star,star points=5,star point ratio=2.3,fill=figGold,draw=figRoute,line width=0.3pt,inner sep=0pt,minimum size=7.5pt]{};}
\DeclareRobustCommand{\legOct}[2]{\tikz[baseline=-0.6ex]\filldraw[fill=#1,draw=#2,line width=0.45pt] (22.5:0.38em) -- (67.5:0.38em) -- (112.5:0.38em) -- (157.5:0.38em) -- (202.5:0.38em) -- (247.5:0.38em) -- (292.5:0.38em) -- (337.5:0.38em) -- cycle;}
\definecolor{tabHead}{gray}{0.92}
\colorlet{tabGroup}{figCoral!10}
\usepackage{microtype}
\makeatletter
\g@addto@macro\normalsize{%
  \setlength\abovedisplayskip{6pt plus 2pt minus 2pt}%
  \setlength\belowdisplayskip{6pt plus 2pt minus 2pt}%
  \setlength\abovedisplayshortskip{2pt plus 2pt}%
  \setlength\belowdisplayshortskip{4pt plus 2pt minus 1pt}}
\makeatother
\AtBeginEnvironment{proof}{\vspace{-4pt}}

\AtEndPreamble{%
  \crefformat{section}{\S#2#1#3}%
  \Crefformat{section}{\S#2#1#3}%
  \crefformat{subsection}{\S#2#1#3}%
  \Crefformat{subsection}{\S#2#1#3}%
  \crefformat{subsubsection}{\S#2#1#3}%
  \Crefformat{subsubsection}{\S#2#1#3}%
  \crefrangeformat{section}{\S\S#3#1#4--#5#2#6}%
  \Crefrangeformat{section}{\S\S#3#1#4--#5#2#6}%
  \crefrangeformat{subsection}{\S\S#3#1#4--#5#2#6}%
  \Crefrangeformat{subsection}{\S\S#3#1#4--#5#2#6}%
  \crefmultiformat{section}{\S\S#2#1#3}{ and #2#1#3}{, #2#1#3}{, and #2#1#3}%
  \Crefmultiformat{section}{\S\S#2#1#3}{ and #2#1#3}{, #2#1#3}{, and #2#1#3}%
  \crefmultiformat{subsection}{\S\S#2#1#3}{ and #2#1#3}{, #2#1#3}{, and #2#1#3}%
  \Crefmultiformat{subsection}{\S\S#2#1#3}{ and #2#1#3}{, #2#1#3}{, and #2#1#3}%
}

\usetikzlibrary{shapes.misc}
\tikzset{
  figfont/.style={font=\footnotesize},
  meas/.style={circle, inner sep=0pt, minimum size=5pt},
  pred/.style={circle, inner sep=0pt, minimum size=5pt, line width=0.8pt, fill=white},
  ring/.style={circle, inner sep=0pt, minimum size=9pt, line width=0.8pt},
  arr/.style={-{Stealth[length=4pt,width=3pt]}, line width=0.9pt},
  parr/.style={-{Stealth[length=4pt,width=3pt]}, line width=0.7pt, densely dashed},
  darr/.style={{Stealth[length=4pt,width=3pt]}-{Stealth[length=4pt,width=3pt]}, line width=0.8pt},
  panel/.style={fill=figPaper!28, rounded corners=4pt},
  ttl/.style={anchor=south west, font=\footnotesize\bfseries, inner sep=0pt},
  note/.style={anchor=south, font=\scriptsize, figMuted, inner sep=0pt},
  lab/.style={inner sep=1.5pt},
  state/.style={circle, draw=figMuted, fill=white, inner sep=0.5pt, minimum size=11pt, font=\scriptsize},
  block/.style={draw, rounded corners=3pt, inner sep=4pt, minimum height=16pt, font=\footnotesize},
}

\DeclareRobustCommand{\legSolidArr}{\tikz[baseline=-0.6ex]\draw[arr, figNavy] (0,0) -- (1.3em,0);}
\DeclareRobustCommand{\legDashedArr}{\tikz[baseline=-0.6ex]\draw[parr, figNavy] (0,0) -- (1.3em,0);}
\DeclareRobustCommand{\legFilled}{\tikz[baseline=-0.6ex]\node[meas, fill=figNavy]{};}
\DeclareRobustCommand{\legHollow}{\tikz[baseline=-0.6ex]\node[pred, draw=figNavy]{};}

\newcommand{\xdocref}[4]{#1~\ifcsname r@#2#3\endcsname\ref{#2#3}\else#4\fi}
\newcommand{\chg}[3]{\textbf{\boldmath #1}~(#2\,$\rightarrow$\,#3)}
\newcommand{\supp}[2]{\xdocref{Supp.}{supp:}{#1}{#2}}
\newcommand{\supptab}[2]{\xdocref{Tab.}{supp:}{#1}{#2}}
\newcommand{\mainref}[3]{\xdocref{#1}{main:}{#2}{#3}}
\newcommand{\asm}[1]{\textsf{A#1}}
\newcommand{\exptabular}[3]{%
  \sbox0{\begin{tabular}{#2}#3\end{tabular}}%
  \setlength{\tabcolsep}{\dimexpr\tabcolsep+(\linewidth-\wd0)/\numexpr2*#1\relax\relax}%
  \begin{tabular}{#2}#3\end{tabular}%
}
\usepackage{needspace}

\usepackage{algorithm}
\usepackage{algpseudocode}
\usepackage{listings}
\definecolor{codecomment}{rgb}{0.42,0.55,0.58}
\lstdefinestyle{pytorch}{language=Python, basicstyle=\fontsize{7.6pt}{8.9pt}\ttfamily\selectfont,
  commentstyle=\color{codecomment}, keywordstyle={}, stringstyle={}, columns=fullflexible,
  keepspaces=true, showstringspaces=false, upquote=true, xleftmargin=1.5em,
  aboveskip=2pt, belowskip=0pt}
\AtEndPreamble{\crefname{algorithm}{Alg.}{Algs.}\Crefname{algorithm}{Algorithm}{Algorithms}}

\newcommand{\supptabsetup}{\centering\footnotesize\renewcommand{\arraystretch}{1.12}\setlength{\aboverulesep}{0pt}\setlength{\belowrulesep}{0pt}\setlength{\tabcolsep}{2pt}}
\newenvironment{pendTable}{%
  \par\addvspace{8pt}\noindent\begin{minipage}{\columnwidth}%
  \captionsetup{type=table}\supptabsetup%
}{\end{minipage}\par\addvspace{8pt}}
\renewcommand{\supp}[2]{\xdocref{Supp.}{}{#1}{#2}}
\renewcommand{\supptab}[2]{\xdocref{Tab.}{}{#1}{#2}}
\renewcommand{\mainref}[3]{\xdocref{#1}{}{#2}{#3}}

\definecolor{cvprblue}{rgb}{0.21,0.49,0.74}
\usepackage[pagebackref,breaklinks,colorlinks,allcolors=cvprblue]{hyperref}

\def\confName{CVPR}
\def\confYear{2027}

\title{Two Actions from One State: Fork Futures Teach\\ JEPA World Models to Compare}

\author{
    Yihao Wu$^{1}$,
    Runjia Zeng$^{2}$,
    Tongyan Wang$^{1}$,
    Renzhong Huang$^{2}$,
    Chongzhou Fang$^{2}$,\\
    Qifan Wang$^{3}$,
    Fan Heng$^{4}$, 
    Dongfang Liu$^{1}$ \\
    $^{1}$Purdue University \quad
    $^{2}$Rochester Institute of Technology \\
    $^{3}$Meta AI \quad
    $^{4}$University of North Texas
}

\begin{document}
\raggedbottom
\maketitle
\begin{abstract}
JEPA world models are trained to predict logged transitions but used to compare counterfactual actions from one state. The logged behavior data show one action per state, and under a local linear approximation we prove the action response is unidentified along any direction the policy never varies within a state. We therefore supplement behavior data with measured outcomes of alternative actions from the same state. Our fork futures execute two actions from one restored state, and their outcome difference cancels the state term, identifying the action response on the sampled span. The resulting model, Fork-JEPA, matches predicted to measured differences, anchors the logged successor with behavior prediction, and adds a lookback head that keeps the predecessor recoverable, all at training time, leaving planning unchanged. On three visual control tasks, behavior-only training leaves the predicted action effect no better than a constant guess. Fork-JEPA makes it readable and raises planning success by up to 14\% over LeWM, its base model. Where the action response changes across hidden zones, it nearly doubles cross-zone success, and 1,024 fork pairs beat 1.2M extra behavior steps. Our project is available \href{https://fork-jepa.github.io/}{here}.


\end{abstract}

\section{Introduction}
\label{sec:intro}
\begin{figure}[t]
\centering
\includegraphics[width=\linewidth]{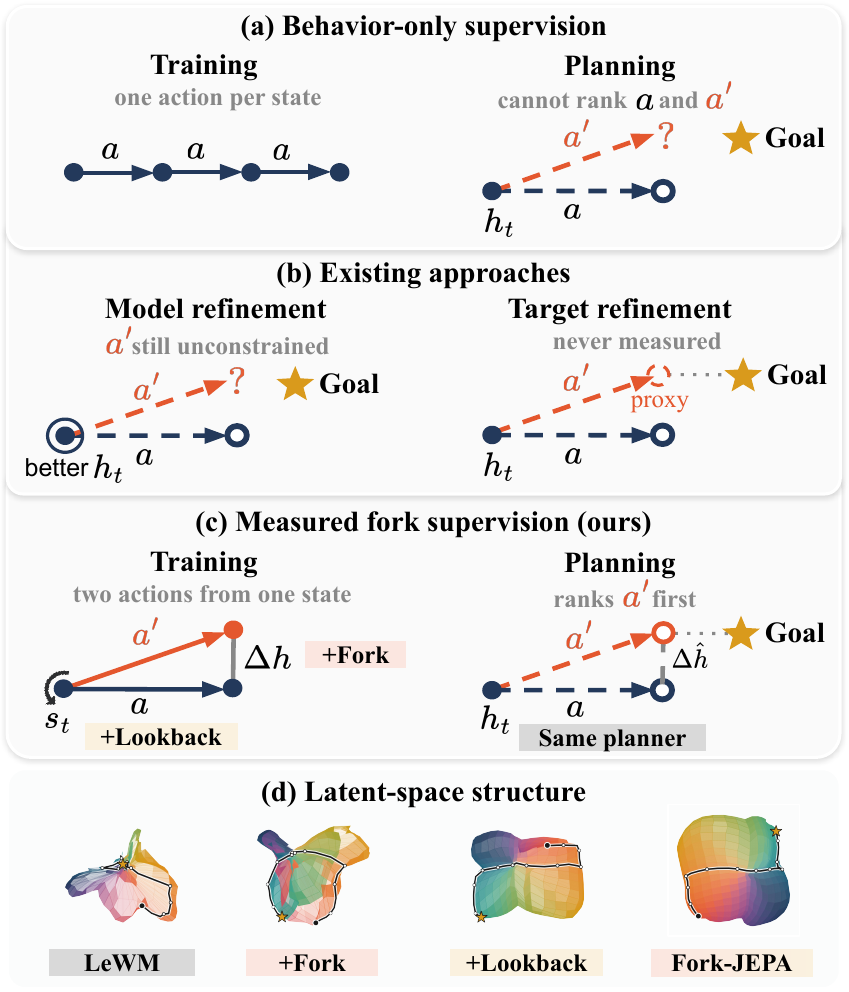}
\vspace{-8pt}
\caption{\textbf{The gap is in the data.} \legSolidArr{} and \legFilled{} are measured, \legDashedArr{} and \legHollow{} predicted. (a) Training fits logged action $a$ at each state (left), leaving $a'$ unconstrained when planning compares their goal distances (right). (b) Model refinement sharpens $h_t$ but not the response to $a'$ (left); synthesized targets remain predictions (right). Neither changes data. (c) Forks restore $s_t$ and execute both actions to measure their outcomes (left). Under the local response model, their difference $\Delta h$ cancels the shared state term and supervises $\Delta\hat h$ (right). Lookback encourages predecessor recovery; planning is unchanged. (d) Isomap latent projections for the same state grid and trajectory in the regime world (\cref{sec:exp-setup}). Lookback promotes state-map continuity; forks supervise action-effect prediction.}
\label{fig:teaser}
\vspace{-4pt}
\end{figure}

Joint-embedding predictive architectures (JEPAs) give world models a latent space to plan in. It predicts the latent outcome of each candidate action and selects the closest to a goal~\cite{lecun2022path, dinowm, vjepa2, leworldmodel}. Each plan thus compares actions at one state, while each logged transition shows one action per state (\cref{fig:teaser}). When the policy barely varies its action at a state, a predictor can fit every logged transition and still know nothing about the alternatives the planner asks about. The planner ranks alternatives using predictions not directly constrained by data (\cref{fig:teaser}a). This raises a fundamental question. \textbf{(Q1)} \emph{How can a JEPA world model learn to compare actions whose outcomes it has never been shown?}


Existing work pulls two levers (\cref{fig:teaser}b). One changes the \emph{model}, with regularizers against collapse~\cite{bardes2022vicreg, lejepa, sobal2025pldm, leworldmodel}, sparse or identifiable latents~\cite{kuang2026lpwm, klindt2026lejepa}, physical-state grounding~\cite{physgrounded}, hierarchical predictors~\cite{zhang2026hjepa}, and test-time adaptation~\cite{wang2026adajepa}. The other changes the \emph{targets} for the actions the data lack, with separation of perturbed branches~\cite{phylatent} and counterfactual transitions synthesized by another model~\cite{buesing2019woulda, coda}. Neither touches the \emph{data}. The same logged transitions can admit different responses to alternative actions, so fitting them more accurately need not resolve the comparison.



The gap is thus in the data, not the model, and we pull the third lever, changing what the data show at each state. \textbf{(Q2)} \emph{What measured data identify the action effect, and how should a JEPA world model be trained on them?}

We answer Q2 through a local analysis of how behavior data constrain action responses. The diagnosis is that behavior actions are a function of the state. Along a direction the policy never varies at a fixed state, the data carry no information about the action response, however many trajectories are logged, and along a rarely varied one they carry little (\cref{prop:design}). A controlled transition is identified only when the action varies at a fixed state~\cite{cwmid}, the excitation that closed-loop system identification has long required~\cite{forssell1999closedloop, ljung1999}. The same diagnosis names the cure. The comparison a planner needs is a counterfactual, two actions at one state, which no trajectory can show. A simulator, however, can restore a state, and then the counterfactual becomes a paired measurement. Two actions from one restored state give two outcomes whose difference cancels the shared state term and identifies the action response along the sampled directions (\cref{prop:fork}, \cref{fig:teaser}c). Actions whose difference lies in the kernel of the local response reach one outcome and form one class. Behavior data leave these classes unresolved, and a fork measures the gap between two of them. Differences alone are not enough. A shared shift of both predictions preserves the difference yet can flip the planner's ranking, which calls for anchoring to logged outcomes, and matching successors alone can merge distinct predecessors into one latent (\cref{fig:teaser}d), which calls for a reverse objective that keeps predecessors recoverable.

We introduce \textbf{Fork-JEPA}, a JEPA world model trained on measured action comparisons. Fork-JEPA restores logged simulator states and re-executes the logged action alongside an alternative, producing \emph{fork futures}, measured outcomes of two actions from one state. A fork-pair loss matches the predicted latent difference to the measured one, behavior prediction anchors the logged successor, and a lookback head reconstructs the predecessor latent from the measured successor and the action. Nothing else changes. The encoder, predictor, and planner are those of LeWM~\cite{leworldmodel}, and forks and the lookback head are used only in training.

Fork-JEPA enjoys three attractive qualities. \textbf{\ding{182} Identified, not guessed, action effects.} Behavior data leave the action response unidentified along any direction the policy never varies at a fixed state (\cref{prop:design}), so a predictor can only guess there, no model change recovers it, and more behavior data help only in proportion to the variation the policy supplies. Forks identify it on the sampled span (\cref{prop:fork}). State readability and action-effect identifiability are separate, and a readable state alone does not make the predictor compare actions correctly (\cref{sec:effect}, \cref{sec:ablations}). \textbf{\ding{183} Measured, not imagined, counterfactuals.} A fork is a paired intervention run in the environment (\cref{sec:data}), so the supervision is an outcome the world produced, not one a model guessed. A prediction shaped by perturbed branches or synthesized transitions is only ever checked against another prediction. To our knowledge, Fork-JEPA is the first JEPA world model trained on executed alternatives. \textbf{\ding{184} A minimal fix.} One state and a second action is the smallest intervention that identifies the action effect. The fork pairs amount to about 0.5\% of the behavior data, and nothing changes at planning time (\cref{sec:objective}).

We verify each claim on TwoRoom~\cite{sobal2025pldm}, PushT~\cite{dinowm}, and OGBench-Cube~\cite{park2025ogbench}. With the same planner, Fork-JEPA improves on LeWM~\cite{leworldmodel} on every task and matches or exceeds, within seed spread, the retrained PLDM~\cite{sobal2025pldm}, DINO-WM~\cite{dinowm}, and goal-conditioned policies~\cite{park2025ogbench} (\cref{sec:planning}). Its predicted action effect is readable where behavior-only training leaves it unreadable (\cref{sec:effect}). In a world whose action response changes across hidden zones, it crosses zones about twice as often as LeWM and ahead of the sparse-latent LpWM~\cite{kuang2026lpwm}, with fork pairs worth 0.5\% of the data outweighing 60\% more behavior data (\cref{sec:regimes}, \cref{sec:ablations}). We believe measured action comparisons offer a principled and practical source of supervision for JEPA world models.
\section{Related Work}
\label{sec:related}
\begin{figure*}[t]
    \centering
    \input{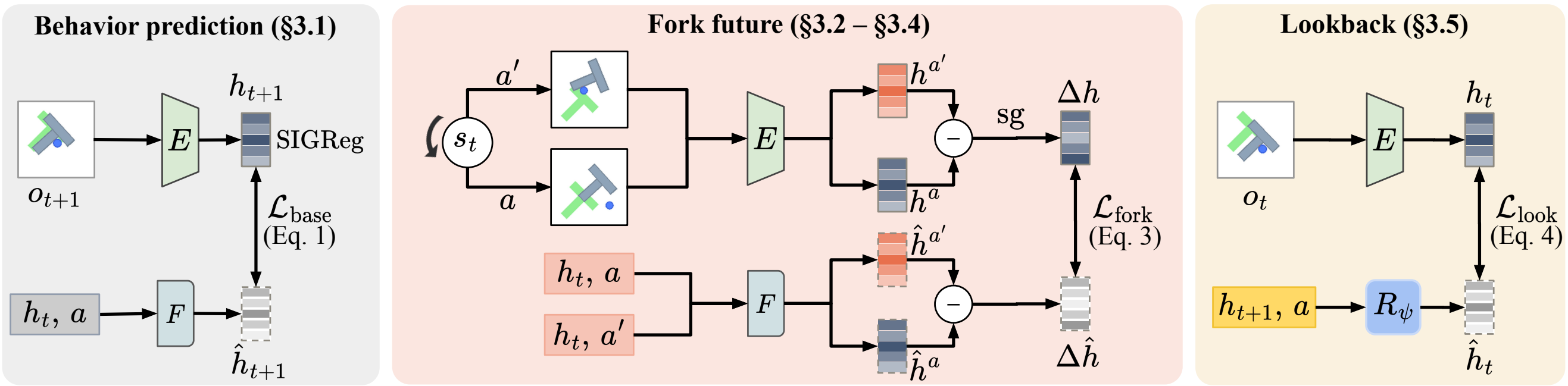}
    \caption{\textbf{Fork-JEPA.} Each loss compares an encoded target with its prediction. \textbf{Left: behavior prediction.} The logged clip supervises successor prediction by $F$. \textbf{Middle: fork future.} A fork executes the logged action $a$ and an alternative $a'$ from the same restored logged state. The encoded outcomes $h^a$ and $h^{a'}$ are compared with their predictions $\hat h^a$ and $\hat h^{a'}$ through their differences: the measured difference cancels the shared state term and provides the stop-gradient (sg) target for the predicted difference. \textbf{Right: lookback.} The reverse head $R_\psi$ reconstructs the predecessor from the measured successor and action. Forks and the reverse head are used only during training.}
    \label{fig:framework}
    \vspace{-4pt}
\end{figure*}
\noindent\textbf{Latent world models and JEPA planning.}
A decade of world models has changed almost everything about the model and almost nothing about its supervision. What they predict ranges over pixels or tokens~\cite{iris, storm, deltairis, diamond, mwm, ivideogpt, dreamer4}, video at foundation scale~\cite{genie, unisim, gaia1, cosmos, longmem, nwm, unipi, dreamgen, uwm, worldvla, genieenv, ctrlworld}, and latent features, as in JEPAs~\cite{lecun2022path, assran2023ijepa, bardes2024vjepa, seqjepa}. They avoid collapse by regularization~\cite{grill2020byol, simsiam, barlow, bardes2022vicreg, dino, dinov2, dimcollapse, lejepa}, freeze the encoder~\cite{dinowm, vjepa2} or train it end to end~\cite{sobal2025pldm, leworldmodel}, and shape the latent to be identifiable~\cite{klindt2026lejepa, zheng2026dsreg}, sparse~\cite{kuang2026lpwm}, or grounded in physical state~\cite{physgrounded, phylatent, yan2026psgjepa, man2026gjepa}. Hierarchy~\cite{director, zhang2026hjepa, masip2026ffjepa} and test-time updates~\cite{wang2026adajepa} extend them, and benchmarks compare them~\cite{whatdrives, swm}. On every axis the predictor is fit to logged transitions, one action per state (\cref{prop:design}), and even a fully grounded latent inherits that limit. Our model is LeWM's. What changes is the data.

\noindent\textbf{Counterfactual supervision and identifiability.}
Faced with a comparison the data lack, prior work changes the targets or nothing. A changed target is a proxy, a margin between predictions for perturbed actions~\cite{phylatent} or a transition synthesized by another model~\cite{buesing2019woulda, coda, mocoda, lu2020cf}, and a proxy is checked only against another prediction. PhyLatent~\cite{phylatent} comes closest. It names the failure and executes two actions from one state to diagnose it, but trains on perturbed predictions, never on the measured pair. Identifiability theory recovers latent causes only with auxiliary variables, sparsity, or interventions~\cite{khemakhem2020ivae, scholkopf, lachapelle, lippe2022citris, icitris, biscuit, brehmer, ahuja, squires, buchholz, vonkugelgen, varici, richens}, and a controlled transition needs the action to vary at a fixed state~\cite{cwmid}, but the variation is taken as given. Without it, offline RL turns pessimistic~\cite{offlinerl, cql, mopo, jinpess, rashidinejad2021bridging, xie2021bellman}, causal effects are only bounded~\cite{koo2026causal, benmichael2025partial}, abstractions drop distractors without adding variation~\cite{cdl, denoisedmdp}, and imitation confuses cause with correlate~\cite{causalconfusion}. None executes the alternative.

\noindent\textbf{Excitation and active system identification.} Closest to our route is control, which designs the comparison. Persistence of excitation~\cite{ljung1999}, identification under feedback~\cite{forssell1999closedloop}, and finite-sample analyses~\cite{dean, sarkar, simchowitz2018learning, ziemann} say how much an input must vary, and experiment design, active exploration, and intrinsically motivated RL choose the variation~\cite{pukelsheim2006optimal, wagenmaker2020active, wagenmaker21, mania, wagenmaker23, lee2024, rnd, plan2explore, memmel2024asid}. All excite a global model along a trajectory. A fork is a paired experiment at one state, where the state term cancels and the action response is identified locally, with no dynamics model in between.


\noindent\textbf{Inverse, backward, and latent-action models.}
Inverse models predict the action between two observations, so the latent keeps what the agent controls~\cite{agrawal2016poke, pathak2017curiosity, efroni, acro, lamb2023acstate, idmpretrain, pclast}, latent-action models infer it from video~\cite{genie, lapo, lapa, igor, univla}, and JEPA world models use inverse heads against collapse~\cite{sobal2025pldm, zhang2026hjepa, physgrounded}. Backward models predict the predecessor~\cite{goyal2019recall}, and invertible networks lose nothing about their input~\cite{jacobsen2018irevnet, behrmann2019invertible}. Our lookback head is a backward map conditioned on the action and applied to the measured successor, which keeps predecessors separable without constraining the architecture, and \cref{sec:ablations} shows that it and the forks add different parts of the gain.
These approaches leave a gap in training: direct supervision from the measured outcomes of different actions executed from the same state. Fork-JEPA fills this gap with fork pairs.
\section{Method}
\label{sec:method}

\subsection{Setup and the Gap}
\label{sec:setup}
\label{sec:design}
An encoder $E$ maps each frame to a latent. A predictor $F$ takes the three latest latents, written $h_t=E(o_{t-2:t})$ for short, together with an action block $a_t\in\mathbb R^m$, and predicts the next frame's latent, $\hat h_{t+1}=F(h_t,a_t)$. Trained on behavior trajectories $\mathcal D_\beta$, the model minimizes:
{
\begin{equation}
    \mathcal L_{\rm base}=\mathbb E_{\mathcal D_\beta}\|\hat h_{t+1}-h_{t+1}\|^2+\lambda_{\rm sig}\,\mathcal L_{\rm SIGReg}(h),
    \label{eq:base}
\end{equation}
}
where SIGReg~\cite{lejepa} keeps the latents from collapsing~\cite{leworldmodel}. At test time, CEM~\cite{rubinstein1999cem} searches for the action sequence whose predicted rollout ends closest to the encoded goal.

\noindent\textbf{Trained to predict, used to compare.} Training matches $\hat h_{t+1}$ to the logged successor. Planning compares. For two actions $a,a'$ from one state, let $h^a,h^{a'}$ be the encoded successors measured in the environment, $\hat h^a,\hat h^{a'}$ their predictions, and $\Delta h=h^a-h^{a'}$, $\Delta\hat h=\hat h^a-\hat h^{a'}$. For generic outcomes $y_a,y_{a'}$ and goal $y_g$ (latents $h^a,h^{a'},h_g$ in JEPA), the planner keeps the candidate with the smaller squared distance. Define the cost gap $D$ as:
{
\begin{equation}
D=\|y_a-y_g\|^2-\|y_{a'}-y_g\|^2.
\label{eq:costdiff}
\end{equation}
}
Thus, $D>0$ favors $a'$, while $D<0$ favors $a$. A predictor can match every logged successor and still get the sign of \cref{eq:costdiff} wrong for $a'$ (\cref{fig:teaser}a).

\noindent\textbf{Why behavior data cannot supply it.} Behavior actions depend on state. Take a locally linear response $y=u(s)+B_aa+\epsilon$ under a feedback policy $a=-Mr(s)+\zeta$ with $\mathbb E[\zeta\mid s]=0$. $u(s)$ is the shared state-dependent term, $B_a$ the action-response matrix, and $\epsilon$ the response residual. The feedback matrix $M$ determines the mean action $-Mr(s)$ from state features $r(s)$; $\zeta$ is the deviation from that mean. The marginal covariance of $a$ can be full rank while $\Sigma_a=\mathbb E_s[\operatorname{Cov}(a\mid s)]$, its fixed-state variation, is small.
\needspace{4\baselineskip}
\begin{proposition}[Conditional Action Design]
\label{prop:design}
For $N$ samples, let $\phi_t=[r(s_t);a_t]$ be the stacked state-action feature vector and $G_N=\sum_t\phi_t\phi_t^\top$ be its unnormalized Gram matrix. For any action-space direction $w$ and the associated joint feature-space direction $v_w=[M^\top w;w]$, $v_w^\top G_Nv_w=\sum_{t=1}^N(w^\top\zeta_t)^2$, whose expectation under a fixed state distribution is $Nw^\top\Sigma_aw$.
\end{proposition}
\begin{proof}[Proof sketch]
Along $v_w$ the state part of the action cancels, $v_w^\top\phi_t=w^\top Mr(s_t)+w^\top a_t=w^\top\zeta_t$, so the quadratic form sums only the squared noise along $w$. See \supp{sec:a1}{A.1}.
\end{proof}


\noindent Under this local model, behavior data constrain observed transitions but may leave alternative-action responses ambiguous (\cref{prop:design}; \supp{sec:a1}{A.1}). More trajectories need not resolve this ambiguity along directions without within-state action variation. This motivates supplementing behavior data with measured alternatives (quality \ding{182} of \cref{sec:intro}). The next three subsections supply the data, its label, and the loss.

\subsection{Fork Futures as Data}
\label{sec:fork}
\label{sec:identify}
\label{sec:data}

A fork supplies the missing action variation through measured outcomes in the environment. From a logged state $s_t$, the fork root, we restore the simulator and execute the logged action $a$ and an alternative $a'$, producing two measured successors as illustrated in \cref{fig:teaser}c. We sample the difference $\delta=a-a'$ isotropically over the full action block, giving each direction the same expected budget. \cref{tab:span} lists the fork budgets $P$, and \cref{sec:span} checks their coverage. Each pair costs two action blocks, so 1,000 pairs add 0.5\% to a 2M-step dataset. Restoration covers the full task state; \supp{sec:b5}{B.2} reports its precision, and \supp{sec:restore-sens}{E.7} tests sensitivity to restoration noise. Under the local response model, the shared state term cancels in the branch difference, whereas different starting states would leave a state difference behind.

\subsection{The Measured Difference as Label}
\label{sec:label}
\begin{figure}[t]
\centering
\includegraphics[width=\linewidth]{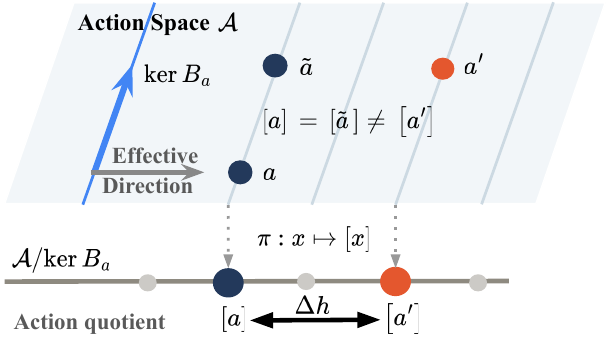}
\caption{\textbf{Action quotient.} At a fixed state under the local response model, $\pi:x\mapsto[x]$ maps each action to its equivalence class $[x]$, the set of actions differing from $x$ only along $\ker B_a$ and sharing its mean response. A fork measures the encoded successor difference $\Delta h=h^a-h^{a'}$ for actions from two classes.}
\label{fig:quotient}
\end{figure}

Encoding the two successors gives the measured difference $\Delta h=h^a-h^{a'}$, in which the shared state term cancels and the action response remains (\cref{fig:teaser}c).
\needspace{4\baselineskip}
\begin{proposition}[Identifiability on the Fork Span]
\label{prop:fork}
Let $\Delta y=y(s,a)-y(s,a')$ and $\Delta\epsilon=\epsilon-\epsilon'$ denote the branch response and residual differences. Suppose the local response $y=u(s)+B_aa+\epsilon$ holds with $\mathbb E[\Delta\epsilon\mid s,a,a']=0$ (\asm{1}), both branches start from the same state (\asm{2}), and the fork differences $\delta=a-a'$ lie in a subspace $\mathcal S$ with orthonormal basis $U$ and positive definite second moment $\mathbb E[zz^\top]$, with coordinates $z=U^\top\delta$ (\asm{3}). Then $\Delta y=B_a\delta+\Delta\epsilon$, and any local linear predictor $\hat y(a)=\hat u+Ba$ with estimated state term $\hat u$ and action-response matrix $B$ that minimizes the population fork-pair loss satisfies $BU=B_aU$.
\end{proposition}
\begin{proof}[Proof sketch]
Subtracting the branches removes $u(s)$, and the loss is $\mathbb E\|(B-B_a)Uz\|^2$ plus a constant (\supp{sec:a2}{A.2}).
\end{proof}
\noindent The action effect is identified on the sampled span and free elsewhere, and if $B_a$ varies with state it holds wherever the local model does. Actions with $a-\tilde a\in\ker B_a$ have the same response and form one class of the quotient (\cref{fig:quotient}). Behavior data leave these classes unresolved. A fork measures the gap between two of them. Nothing is fitted between the branches, so the label needs no dynamics model, unlike excitation along a trajectory.

\subsection{The Fork-Pair Loss}
\label{sec:loss}
The second half of Q2 is how to train on it. We train the predicted difference to match the measured one:
{
\begin{equation}
    \mathcal L_{\rm fork}
    =\mathbb E_{\rm pair}\big\|\Delta\hat h-\operatorname{sg}(\Delta h)\big\|^2,
    \label{eq:lca}
\end{equation}
}
where $\operatorname{sg}$ stops the gradient through the measured target~\cite{grill2020byol}. The target is a measurement, not another prediction, so the loss cannot be met by a predictor that agrees with itself, which separates it from a margin between perturbed predictions or a synthesized transition. Regressing each branch onto its own successor would also use the pair, but each residual then carries the shared state error, which dominates the small action effect and cancels in the difference.

\noindent\textbf{Anchoring the pair.} Writing \(\bar y=(y_a+y_{a'})/2\), the cost gap in Eq. (2) is \(2\langle y_a-y_{a'},\bar y-y_g\rangle\). A common shift of both predictions leaves their difference and \(\mathcal L_{\mathrm{fork}}\) unchanged, but moves the midpoint relative to the goal and can reverse the ranking. Behavior prediction anchors the logged successor. With transition errors $e(a)=\hat h^a-h^a$, the fork residual is $e(a)-e(a')$, so $\|e(a')\|\le\|e(a)\|+\|\Delta\hat h-\Delta h\|$, and an accurate logged successor and a small fork residual together bound the error of the alternative.

\subsection{Behavior Lookback}
\label{sec:lookback}
Both losses act on successors and can be satisfied even when distinct predecessors map to the same successor latent under the same action. SIGReg constrains the latent distribution but does not ensure predecessor recoverability. To preserve predecessor information, we add a reverse head $R_\psi$, with trainable parameters $\psi$, called lookback, that reconstructs the predecessor from the measured successor and action:
{
\begin{equation}
    \mathcal L_{\rm look}=\mathbb E_{\mathcal D_\beta}\big\|R_\psi(h_{t+1},a_t)-h_t\big\|^2.
    \label{eq:llog}
\end{equation}
}
If both reconstruction errors are below $\varepsilon$ and $R_\psi$ is $\ell$-Lipschitz, two predecessors under the same action have measured successors at least $(\|h_t-h_t'\|-2\varepsilon)/\ell$ apart, so predecessors more than $2\varepsilon$ apart cannot share one. The gradient reaches the shared encoder, encouraging information retention. Lookback needs no environment calls, while forks exist only at selected states. State readability and action comparability are the two halves of quality \ding{182}, and \cref{sec:ablations} shows that lookback supports the former and forks the latter.


\subsection{Objective and Planning}
\label{sec:objective}
\label{sec:analysis}
Fork-JEPA minimizes the three terms jointly (\cref{fig:framework}):
{
\begin{equation}
    \mathcal L=\mathcal L_{\rm base}+\lambda_{\rm fork}\mathcal L_{\rm fork}+\lambda_{\rm look}\mathcal L_{\rm look},
    \label{eq:objective}
\end{equation}
}
updating the encoder, predictor, and reverse head together. Each step draws a behavior batch for the first and last terms and a fork batch for the second. Nothing else changes. The reverse head and the forks exist only in training, so inference costs what it did before (quality \ding{184}).
\needspace{4\baselineskip}
\begin{proposition}[Planning Regret]
\label{prop:regret}

In a deterministic, fully observed setting with an isometric reference embedding (\supp{sec:a3}{A.3}), consider true and predicted squared goal-distance costs $c,\hat c$ over a finite set $\mathcal C$ of length-$H$ action sequences. Let the encoding errors at the start and the goal be at most $\varepsilon_{\rm rep}$, each candidate's rollout error at most $\varepsilon_{\rm tr}(\mathbf a)$, $F$ be $L$-Lipschitz in its latent input, and the true and predicted terminal-to-goal vectors have norm at most $R$. Then the planner's choice $\hat{\mathbf a}\in\arg\min_{\mathcal C}\hat c$ satisfies:
{
\begin{equation}
    c(\hat{\mathbf a})-\min_{\mathcal C}c
    \le4R\big[(1+L^H)\varepsilon_{\rm rep}
    +\max_{\mathcal C}\varepsilon_{\rm tr}(\mathbf a)\big].
\label{eq:regret}
\end{equation}
}
\end{proposition}
\begin{proof}[Proof sketch]
Each candidate's predicted and true costs differ by at most $2R[(1+L^H)\varepsilon_{\rm rep}+\varepsilon_{\rm tr}(\mathbf a)]$, and comparing the chosen with the best candidate doubles it (\supp{sec:a3}{A.3}).
\end{proof}
\noindent The transition term is a maximum over all candidates, so a predictor accurate only on logged actions can still pick the wrong plan. Forks and behavior prediction together bound the error on the alternatives. The representation term is amplified by $L^H$ over the rollout, so the latent must keep the state, which SIGReg and lookback maintain. \cref{sec:ablations} tests which term supplies which part.

\section{Experiments}
\label{sec:exp}

\cref{sec:exp-setup} gives the setup. \cref{sec:effect} shows that a predictor trained on behavior data fits the next state yet loses the action effect, and that forks restore it (Q1, quality \ding{182}). \cref{sec:planning} shows that the restored comparison carries over to planning at unchanged planning cost (quality \ding{184}), \cref{sec:regimes} tests a world whose action response changes with the state, \cref{sec:levers} compares the two levers of \cref{sec:intro} with and without our data (quality \ding{183}), and \cref{sec:ablations} asks which term supplies what, the two halves of quality \ding{182}. The appendix gives proofs, implementation details, and further results and controls.

\subsection{Setup}
\label{sec:exp-setup}

\begin{table}[t]
\centering
\footnotesize
\renewcommand{\arraystretch}{1.12}
\setlength{\aboverulesep}{0pt}
\setlength{\belowrulesep}{0pt}
\setlength{\tabcolsep}{2pt}
\exptabular{6}{llcccr}{
\toprule
\rowcolor{tabHead}
\textbf{Task}\rule{0pt}{2.6ex} & \textbf{Action} ($d_a$) & $m$ & $P$ & $P/m$ & \textbf{Steps}\\
\midrule
TwoRoom & velocity (2) & 10 & 1,000 & 100 & 0.92M\\
PushT   & motion (2)   & 10 & 2,000 & 200 & 2.34M\\
Cube    & effector, gripper (5) & 25 & 1,000 & 40 & 2.00M\\
\bottomrule
}
\caption{\textbf{Tasks, action blocks, and fork budgets.} $m$: five-step action-block dimension; $P$: fork pairs; Steps: behavior-data size in environment steps. $P/m$ gives the average budget per direction under isotropic sampling.}
\label{tab:span}
\end{table}

\noindent\textbf{Environments and data.} We use TwoRoom (2D navigation)~\cite{sobal2025pldm}, PushT (planar pushing)~\cite{dinowm}, and OGBench-Cube (3D manipulation)~\cite{park2025ogbench}, with a noisy heuristic policy for TwoRoom and expert data for PushT and Cube. Dataset sizes and fork budgets are in \cref{tab:span}. We adapt LpWM's Piecewise world~\cite{kuang2026lpwm} into a unit-square regime world with four hidden zones, replacing zone-dependent drift with zone-dependent action matrices $B_z$ (\cref{fig:regimemaps}; \supp{sec:regime-world}{E.1}).

\noindent\textbf{Baselines.} We compare with LeWM~\cite{leworldmodel}, PLDM~\cite{sobal2025pldm}, DINO-WM~\cite{dinowm}, and goal-conditioned policies GCBC, GCIQL, and GCIVL~\cite{park2025ogbench}. The regime world adds LpWM~\cite{kuang2026lpwm}, with and without temporal Jaccard (TJ), and \cref{tab:paradigms} carries PhyLatent's reported results~\cite{phylatent}. GA-WM~\cite{physgrounded} uses privileged physical-state supervision, so it enters only \cref{tab:paradigms}, not the image-only comparison of \cref{fig:sr}.

\noindent\textbf{Training and evaluation.} Fork-JEPA uses a ViT-Tiny encoder~\cite{dosovitskiy2021vit} trained from scratch for 10 epochs with AdamW, batch size 128, and weight decay $10^{-3}$. Task-specific settings are in \supptab{tab:config}{S2}. Visual-task baselines are retrained in the same pipeline and evaluated on 50 shared development episodes over three seeds. World models use matched CEM planning~\cite{rubinstein1999cem, swm} over five five-step action blocks (\supp{sec:b3}{B.3}); goal-conditioned policies act directly. Regime-world models share 10,000 behavior episodes, architecture, initialization, and training budget. Fork-JEPA additionally uses 1,024 fork pairs and lookback; LeWM retains its original objective, and LpWM uses RDMReg with Piecewise settings. Evaluation uses 200 fixed start--goal pairs, half cross-zone, with replanning after each block.

\noindent\textbf{Metrics.} We report success rate (SR) with error bars on visual tasks, and cross-zone and overall SR at the horizons in \cref{tab:regimes}. Position $R^2$ and zone balanced accuracy (BA) use linear readouts of frozen latents. Worst-zone predicted action-map error measures regime-response accuracy and restoration sensitivity. Data-efficiency and restoration tests use cross-zone SR with planning horizon $H=10$.


\noindent\underline{\textit{Action-effect probes.}} \MID{} and \MIDD{} measure state and action-effect prediction at rollout depth $k$, using branches that differ only in their first action block:
{
\begin{equation}
\begin{aligned}
\mMID_k&=R^2_{\rm test}(x_k,g_k(\hat h_k)),\\
\mMID_{\Delta,k}&=R^2_{\rm test}(\Delta x_k,g_k^\Delta(\Delta\hat h_k)).
\end{aligned}
\end{equation}
}
Here $g_k$ is a ridge readout~\cite{alain2016probes} fitted on predicted latents; $g_k^\Delta$ is fitted on measured differences and frozen. We split 512 held-out fork roots equally to fit and test, averaging over state dimensions and depths 1--5. Unlike LeWM's frame-level probes~\cite{leworldmodel}, these evaluate full-state readout after off-policy blocks, including Cube velocities and orientations.

\subsection{Predicting Well Is Not Comparing Well}
\label{sec:diagnosing}
\label{sec:effect}
\label{sec:probe}
\label{sec:span}

\noindent\textbf{Fork coverage.} Executed differences span the action space on all tasks, with the smallest eigenvalue of $\sum\delta\delta^\top$ at least 65\% of the mean (\supptab{tab:coverage}{S4}). This supports action-direction coverage in \asm{3} of \cref{prop:fork}, but not coverage of every state-dependent response. We test sensitivity to fork budget and magnitude, and whether gains are confined to sampled action directions. Cube has only 40 pairs per direction (\cref{tab:span}); TwoRoom levels off near $P=1{,}024$, while PushT still improves at $P=2{,}048$ (\supptab{tab:budget-sweep}{S18}). \MIDD{} peaks at the default fork magnitude on both tasks (\supptab{tab:fork-magnitude}{S19}). Subspace-restricted forks yield 0.64 on $\mathcal S$ but 0.10 on $\mathcal S^\perp$, versus about 0.68 on both with isotropic forks (\supptab{tab:span-ablation}{S25}).


\noindent\textbf{Action-effect prediction.} As shown in \cref{tab:probe}. On TwoRoom, LeWM obtains a high state readout score (\MID{} 0.91), but its action-effect score is $-0.10$. Fork-JEPA raises \MIDD{} to 0.69, a gain of 0.79, while improving \MID{} to 0.98. We also observe higher \MIDD{} on PushT (0.08 to 0.30) and Cube (0.00 to 0.06), accompanied by higher \MID{}. Thus, the action-effect gains do not come at the expense of state readout accuracy. The smaller gain on Cube may reflect its more complex contact-dependent responses and lower fork budget per direction.

\begin{table}[t]
\centering
\footnotesize
\renewcommand{\arraystretch}{1.12}
\setlength{\aboverulesep}{0pt}
\setlength{\belowrulesep}{0pt}
\setlength{\tabcolsep}{2pt}
\exptabular{6}{llcccc}{
\toprule
\rowcolor{tabHead}
& & \multicolumn{2}{c}{\textbf{Encoder (measured)}} & \multicolumn{2}{c}{\textbf{Predictor}}\\
\cmidrule(lr){3-4}\cmidrule(l){5-6}
\rowcolor{tabHead}
\multirow{-2}{*}{\textbf{Task}} & \multirow{-2}{*}{\textbf{Model}} & Pos. $R^2$ & \MIDD{} & \MID{} & \MIDD{}\\
\midrule
TwoRoom & LeWM      & 0.992 & 0.47 & $0.91{\scriptstyle\pm0.03}$ & $-0.10{\scriptstyle\pm0.12}$\\
        & Fork-JEPA & 1.000 & \textbf{0.99} & $\mathbf{0.98}{\scriptstyle\pm0.02}$ & $\mathbf{0.69}{\scriptstyle\pm0.07}$\\
\midrule
PushT   & LeWM      & 0.956 & 0.06 & $0.48{\scriptstyle\pm0.08}$ & $0.08{\scriptstyle\pm0.16}$\\
        & Fork-JEPA & 0.993 & \textbf{0.28} & $\mathbf{0.58}{\scriptstyle\pm0.06}$ & $\mathbf{0.30}{\scriptstyle\pm0.10}$\\
\midrule
Cube    & LeWM      & 0.979 & 0.01 & $0.13{\scriptstyle\pm0.07}$ & $0.00{\scriptstyle\pm0.10}$\\
        & Fork-JEPA & 0.996 & \textbf{0.07} & $\mathbf{0.27}{\scriptstyle\pm0.05}$ & $\mathbf{0.06}{\scriptstyle\pm0.06}$\\
\bottomrule
}
\caption{\textbf{State and action-effect readouts.} Readouts on measured and predicted latents. Encoder: position $R^2$ and measured-branch \MIDD{}. Predictor: \MID{} and \MIDD{} over rollout depths 1--5.}
\label{tab:probe}
\vspace{-4pt}
\end{table}

\noindent\textbf{Encoder and predictor analysis.} We compare measured and predicted readouts to locate the action-effect deficit using the probe-transfer bound (\supp{sec:b4}{A.4}). Both encoders achieve controlled-body position $R^2\geq0.95$ on every task (\cref{tab:probe}), and a two-layer MLP preserves the readout ordering (\supptab{tab:mlp-readout}{S6}). On TwoRoom, LeWM's \MIDD{} falls from 0.47 on measured differences to $-0.10$ on predicted ones, indicating a predictor-side deficit. On PushT and Cube, even measured differences score poorly (0.06 and 0.01), suggesting poor linear readability of the encoded action effect. Fork-JEPA improves both measured and predicted scores on all three tasks, consistent with fork-pair supervision of action differences (\cref{sec:loss}). The frozen-encoder control isolates the benefit of predictor training (\cref{tab:frozen}). The bound holds in all 1,000 bootstrap draws on PushT and Cube (\supptab{tab:probedecomp}{S5}).

\begin{figure}[t]
\centering
\includegraphics[width=\linewidth]{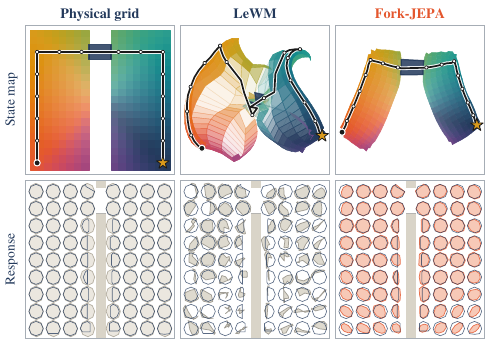}
\caption{\textbf{Latent space and action responses on TwoRoom.} State maps: t-SNE embeddings colored by position. State map: start \legWalkStart{}, doorway \legBox{figNavy}, goal \legWalkGoal{}. Response maps: measured \legOct{white}{figNavy}, action-only \legOct{figSandA}{figSandC}, LeWM \legOct{figLight}{figMuted}, Fork-JEPA \legOct{figCoralLight}{figCoral}.}
\label{fig:tworoom}
\vspace{-4pt}
\end{figure}

\noindent\textbf{Latent-space structure.} We visualize TwoRoom latents with t-SNE and trace a walk through the doorway in \cref{fig:tworoom}. Fork-JEPA forms two coherent regions connected at the doorway, and the walk remains smooth across rooms. LeWM preserves neighborhoods but shows fragmentation and jumps along the longer path. This suggests that Fork-JEPA preserves spatial continuity beyond local neighborhoods, consistent with lookback encouraging recovery of distinct predecessors that successor prediction alone may merge (\cref{sec:lookback}). The spatial readout tests agree, since LeWM's error grows as fitting and evaluation locations move apart while Fork-JEPA's stays low (\supp{sec:wall-grid}{C.3}).

\noindent\textbf{State-dependent action responses.} We plot eight-direction push responses at each free TwoRoom cell in \cref{fig:tworoom}. Closer agreement between predicted and measured octagons indicates greater accuracy. Measured octagons are regular in open space and flatten where walls block motion. Fork-JEPA captures this flattening, while LeWM is less accurate and the action-only readout remains unchanged across cells. This suggests that Fork-JEPA captures state-dependent action effects, consistent with same-state fork supervision in \cref{sec:loss}. The wall-grid evaluation in \supptab{tab:wall-grid}{S8} agrees, with median predicted response error of 0.029 for Fork-JEPA versus 1.10 for LeWM. A violation-of-expectation test provides further support. On TwoRoom, Fork-JEPA is far more surprised by object teleportation than recoloring, while LeWM responds similarly to both, as shown in \supp{sec:voe-supp}{F}. The \MIDD{} gain persists across rollout depths 1--5, as shown in \supp{sec:more-depth}{G.1}.

\subsection{Comparing Well Gives Better Plans}
\label{sec:planning}

\begin{figure}[t]
\centering
\includegraphics[width=\linewidth]{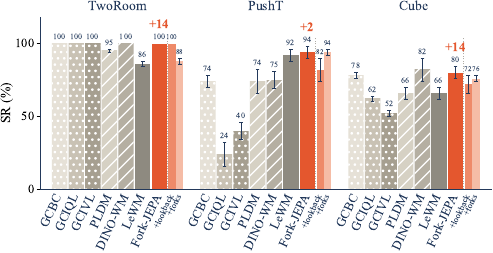}
\caption{\textbf{Planning performance on visual control tasks.} SR (\%) on three visual control tasks. Policies: GCBC \legDots{figSandA}, GCIQL \legDots{figSandB}, GCIVL \legDots{figSandC}. World models: PLDM \legHatch{figLight}, DINO-WM \legHatch{figMid}, LeWM \legBox{figMuted}, Fork-JEPA \legBox{figCoral}. Split bars: +lookback \legBox{figCoralMid}, +forks \legBox{figCoralLight}. Error bars indicate standard deviations over three seeds. 
}
\label{fig:sr}
\end{figure}

\begin{figure}[t]
    \centering
    \includegraphics[width=\linewidth]{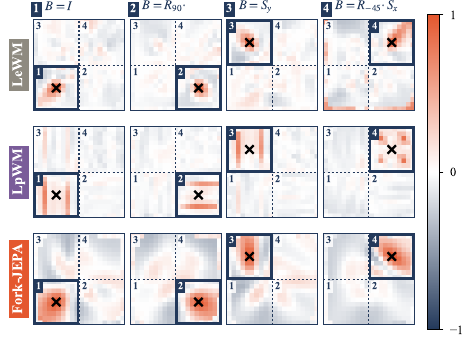}
    \caption{\textbf{Representation maps in the regime world.} Colors indicate cosine similarity to the reference cell $\times$ on a $20\times20$ grid.}
    \label{fig:regimemaps}
\end{figure}

\noindent\textbf{Visual control tasks.} Fork-JEPA outperforms LeWM and PLDM on all three tasks (\cref{fig:sr}). SR is averaged over three seeds on 50 development episodes (\supptab{tab:sr-values}{S8}). Over LeWM, SR improves by 14 percentage points on TwoRoom ($86\pm2$\% to $100\pm0$\%) and Cube ($66\pm4$\% to $80\pm4$\%), but only 2 on PushT (92\% to 94\%), where LeWM already performs well. On Cube, Fork-JEPA is comparable to DINO-WM within seed variation ($80\pm4$\% versus $82\pm8$\%). Remaining failures may reflect how action effects depend on gripper position and contact. The same motion may move only the gripper or also displace the cube. Limited fork coverage of these configurations may leave some planning comparisons weakly constrained. On TwoRoom, lookback alone and Fork-JEPA both reach 100\% SR; response maps and errors assess action-effect prediction more finely (\cref{fig:tworoom}; \supptab{tab:wall-grid}{S8}).

\begin{table}[t]
\centering
\footnotesize
\renewcommand{\arraystretch}{1.12}
\setlength{\aboverulesep}{0pt}
\setlength{\belowrulesep}{0pt}
\setlength{\tabcolsep}{0.7pt}
\exptabular{9}{lcccccccc}{
\toprule
\rowcolor{tabHead}
& \multicolumn{4}{c}{\textbf{Cross-zone SR (\%)}} & \multicolumn{2}{c}{\textbf{SR (\%)}} & \textbf{Pos.} & \textbf{Zone (\%)}\\
\cmidrule(lr){2-5}\cmidrule(l){6-7}
\rowcolor{tabHead}
\multirow{-2}{*}{\textbf{Model}} & $H=5$ & $10$ & $15$ & $20$ & $H=5$ & $10$ & $R^2$ & BA\\
\midrule
LeWM           & 25 & 35 & 30 & 29 & 54.5 & 57.5 & 0.69 & 67.0\\
\hspace{.5em}+lookback       & 60 & 61 & 51 & 36 & 77.0 & 76.5 & 0.86 & 95.4\\
\hspace{.5em}+forks (diff)    & 44 & 48 & 36 & 31 & 70.0 & 70.5 & 0.75 & 90.2\\
\hspace{.5em}+forks (direct)  & 39 & 42 & 32 & 30 & 63.5 & 67.5 & 0.71 & 89.6\\
LpWM           & 44 & 41 & 35 & 28 & 70.5 & 64.5 & 0.87 & \textbf{96.6}\\
LpWM+TJ        & 50 & 61 & \textbf{54} & 33 & 73.0 & 71.5 & 0.89 & 91.4 \\
\midrule
Fork-JEPA (direct)  & 60 & 67 & 51 & 36 & 75.5 & 73.0 & 0.89 & 92.4\\
Fork-JEPA (ours)  & \textbf{61} & \textbf{68} & 52 & \textbf{37} & \textbf{77.5} & \textbf{78.5} & \textbf{0.92} & 88.4\\
\bottomrule
}
\caption{\textbf{Planning and component ablations in the regime world.} (diff): matching branch differences; (direct): regressing each branch separately. Results are means over three seeds.}
\label{tab:regimes}
\vspace{-4pt}
\end{table}

\noindent\textbf{Candidate comparisons.} In \supptab{tab:ranking}{S10}, Fork-JEPA improves latent-cost rankings on all tasks with non-overlapping 95\% bootstrap intervals, but physical-distance rankings improve only on TwoRoom and PushT, revealing a latent--physical cost gap on Cube. In a complementary test on held-out fork pairs, Fork-JEPA selects the physically closer branch more accurately than LeWM at every tested depth (\supp{sec:todo-mechanism}{H.3}).

\noindent\textbf{Computational cost.} In \supptab{tab:cost}{S28}, PushT training time increases from 6.045 to 9.646 hours ($+59.6\%$), with similar CPU planning time (24.090 versus 24.352 s per episode), and the interaction count rises from 2.34M to 2.36M steps.

\subsection{Planning Across Dynamics Regimes}
\label{sec:regimes}
\label{sec:efficiency}

\noindent\textbf{Cross-zone planning.} We report cross-zone planning results in \cref{tab:regimes}. Fork-JEPA achieves 68\% SR at $H=10$, outperforming LeWM, LpWM, and LpWM+TJ by 33, 27, and 7 \%, respectively. It also achieves the highest SR at three of the four horizons. LpWM separates the zones better, with 96.6\% zone accuracy against 88.4\%, yet plans worse, so recognizing a regime is not predicting its action effects, which is what forks supervise. In \cref{tab:paradigms}, Fork-JEPA also has the lowest worst-zone action-map error, below LeWM and LpWM. \supp{sec:regime-examples}{E.5} shows example plans.

\begin{figure}[t]
\centering
\includegraphics[width=\linewidth]{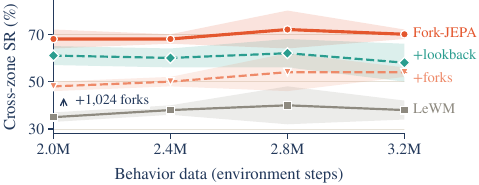}
\caption{\textbf{Data efficiency in the regime world.} Cross-zone SR at $H=10$, with mean and spread over three seeds. The arrow marks the gain of +forks over LeWM at 2.0M behavior steps.}
\label{fig:data-eff}
\vspace{-4pt}
\end{figure}

\noindent\textbf{Data efficiency.} In \cref{fig:data-eff}, adding 60\% more behavior data raises LeWM's cross-zone SR from 35\% to 38\%; forks costing about 0.5\% additional interaction raise it to 48\% at the original behavior budget. On PushT, 60\% more behavior data raises lookback SR from 82\% to 88\%, while measured forks at the original behavior budget reach 94\% (\supptab{tab:data-vs-forks}{S21}).



\noindent\textbf{Restoration robustness.} In \supptab{tab:restore-sens-regime}{S15}, with starting-state offsets up to 8.7 times the median action effect, cross-zone SR stays within 7 \% of exact restoration, while worst-zone map error rises from 0.22 to 6.66. State mismatch weakens the shared-state cancellation in \cref{prop:fork}, degrading action-response accuracy. Planning remains more tolerant, opening the possibility of real-world robot deployment with approximate state restoration. On the visual tasks, restoration noise up to $\sigma = 10^{-2}$ keeps SR within 6~\% and \MIDD{} within 0.027 of exact restoration (\supptab{tab:restore-sens}{S17}).


\subsection{Two Levers versus the Data Lever}
\label{sec:levers}

The introduction groups prior answers to Q1 into two levers, changing the model and changing the targets (\cref{fig:teaser}b), and argues that neither touches the data. \cref{tab:paradigms} tests both claims, running each method as published and again with our fork pairs and fork-pair loss added.

\noindent\textbf{Complementarity of fork supervision.} As shown in \cref{tab:paradigms}, adding forks improves planning across model designs, including Cube SR gains of 8 and 3 percentage points for PLDM and LpWM. For GA-WM, which uses privileged physical-state supervision, forks reduce worst-zone map error from 0.577 to 0.534 and raise cross-zone SR from 71\% to 73\%, showing additional benefits even with physical-state grounding. Measured rather than imagined targets raise TwoRoom \MIDD{} from 0.28 to 0.69. The benefit also extends to RDMReg's sparse latents, where adding forks to lookback raises cross-zone SR from 48\% to 64\% in \supptab{tab:ablate-grid}{S13}. Furthermore, measured forks outperform state-alignment and inverse-dynamics proxies on TwoRoom and PushT action-effect readouts in \supptab{tab:proxy-baselines}{S26}.

\begin{table}[t]
\centering
\footnotesize
\renewcommand{\arraystretch}{1.08}
\setlength{\tabcolsep}{0.7pt}
\setlength{\arrayrulewidth}{0.35pt}
\setlength{\doublerulesep}{1pt}
\exptabular{7}{l||cc|cc|cc}{
\hline\hline
\rowcolor{gray!15}
& \multicolumn{2}{c|}{\textbf{TwoRoom}}
& \multicolumn{2}{c|}{\textbf{Cube}}
& \multicolumn{2}{c}{\textbf{Regime world}}\\
\cline{2-7}
\rowcolor{gray!15}
\multirow{-2}{*}{\textbf{Model}}
& \MIDD{} & SR (\%)
& \MIDD{} & SR (\%)
& Map $\downarrow$ & SR (\%)\\
\hline\hline
\rowcolor{gray!8}
\multicolumn{7}{c}{\textit{Behavior data only}}\\
\hline
LeWM & $-0.10$ & 86 & 0.00 & 66 & 0.739 & 35\\
\hline
\rowcolor{gray!8}
\multicolumn{7}{c}{\textit{Change the model}}\\
\hline
PLDM                & 0.61 & 95  & 0.03 & 66 & 0.277 & 58\\
\quad+forks         & 0.70 & 99  & 0.02 & 74 & 0.254 & 62\\
LpWM                & 0.58 & 96  & 0.01 & 69 & 0.261 & 41\\
\quad+forks         & 0.63 & 98  & 0.01 & 72 & 0.246 & 52\\
GA-WM$^\dagger$     & 0.66 & 98  & 0.02 & 74 & 0.577 & 71\\
\quad+forks         & 0.69 & 100 & 0.04 & 78 & 0.534 & 73\\
AdaJEPA$^\ddagger$  & 0.55 & 94  & 0.02 & 74 & 0.728 & 42\\
\quad+forks         & 0.58 & 100 & 0.03 & 78 & 0.572 & 54\\
\hline
\rowcolor{gray!8}
\multicolumn{7}{c}{\textit{Change the targets}}\\
\hline
PhyLatent$^*$  & --   & 97 & --   & 82 & --    & --\\
Imagined forks & 0.28 & 96 & 0.02 & 70 & 1.034 & 54\\
\hline
\rowcolor{gray!8}
\multicolumn{7}{c}{\textit{Change the data (ours)}}\\
\hline
Fork-JEPA & \textbf{0.69} & \textbf{100}
          & \textbf{0.06} & \textbf{80}
          & \textbf{0.221} & \textbf{68}\\
\hline\hline
}
\caption{\textbf{Fork supervision across model and target designs.} +forks adds measured fork pairs and the fork-pair loss to each method. \MIDD{} is averaged over depths 1--5; Map is worst-zone predicted action-map error; regime SR is cross-zone success at $H=10$. $^\dagger$Trains with privileged physical-state labels. $^\ddagger$Test-time adaptation. $^*$Originally reported results.}
\label{tab:paradigms}
\end{table}

\subsection{Which Term Supplies What?}
\label{sec:ablations}
\label{sec:component-ablations}

\noindent\textbf{Task-dependent contributions of lookback and forks.} In \cref{fig:sr}, lookback raises TwoRoom SR from 86\% to 100\%, while forks alone reach 88\%. Complementarity is clearest on Cube, where lookback and forks individually raise SR from 66\% to 72\% and 76\%, and together reach 80\%. On PushT, lookback lowers SR from 92\% to 82\% and block-orientation readout $R^2$ from 0.81 to 0.49, suggesting reduced access to orientation cues for choosing effective pushes. Forks alone or combined reach 94\%, within LeWM's seed variation. In the regime world, lookback raises position $R^2$ from 0.69 to 0.86 and cross-zone SR from 35\% to 61\% in \cref{tab:regimes}. Adding forks raises SR to 68\% despite reducing zone BA from 95.4\% to 88.4\%, showing that better regime recognition does not necessarily yield better planning. Forks alone raise SR from 35\% to 48\%. The latent maps in \cref{fig:teaser}d illustrate improved state-map continuity with lookback, while \supptab{tab:response}{S12} shows that lookback worsens predicted action-map error, which adding forks reduces from 0.890 to 0.087. Predecessor-retrieval controls in \supp{sec:todo-mechanism}{H.3}  support lookback's recoverability role. Finally, the sweeps in \supptab{tab:loss-weights}{S20} show that PushT SR is more sensitive to $\lambda_{\rm look}$ than to $\lambda_{\rm fork}$.


\begin{table}[t]
\centering
\footnotesize
\renewcommand{\arraystretch}{1.08}
\setlength{\tabcolsep}{2pt}
\setlength{\arrayrulewidth}{0.35pt}
\setlength{\doublerulesep}{1pt}
\exptabular{5}{l||cc|c|c}{
\hline\hline
\rowcolor{gray!15}
& \multicolumn{2}{c|}{\textbf{TwoRoom}}
& \textbf{PushT} & \textbf{Cube}\\
\cline{2-5}
\rowcolor{gray!15}
\multirow{-2}{*}{\textbf{Model}}
& \MIDD{} & SR (\%) & SR (\%) & SR (\%)\\
\hline\hline
\rowcolor{gray!8}
\multicolumn{5}{c}{\textit{Frozen DINOv2 encoder}}\\
\hline
DINO-WM    & 0.26 & 100 & 75 & 82\\
\quad+forks & 0.71 & 100 & 90 & 88\\
\hline
\rowcolor{gray!8}
\multicolumn{5}{c}{\textit{Frozen LeWM encoder}}\\
\hline
LeWM, predictor retrained & 0.01 & 86 & 92 & 66\\
\quad+forks               & 0.21 & 94 & 94 & 72\\
\hline\hline
}
\caption{\textbf{Fork supervision with frozen encoders.} Only the predictor is trained. \MIDD{} is reported on TwoRoom over depths 1--5.}
\label{tab:frozen}
\end{table}

\noindent\textbf{Fork supervision design.} Paired transitions yield similar planning SR but lower map error than independent transitions in \supp{sec:response}{E.2}, supporting same-state measurement as motivated by \cref{prop:fork}. For the training target, difference matching and direct branch regression reach cross-zone SR of 48\% and 42\% without lookback in \cref{tab:regimes}, both above LeWM's 35\%. With lookback, their scores are similar at 68\% and 67\%. Measured alternatives therefore help with either target. Difference matching helps more without lookback, possibly because it cancels shared state error when the representation is less constrained, as discussed in \cref{sec:loss}. It also outperforms direct regression on all three visual tasks in \supptab{tab:direct-vs-diff}{S11}. Finally, held-out residual audits in \supptab{tab:logged-vs-alt}{S23} confirm the anchoring bound on every task.

\noindent\textbf{Frozen-encoder analysis.} In \cref{tab:frozen}, forks raise TwoRoom \MIDD{} from 0.01 to 0.21 with LeWM's encoder fixed and from 0.26 to 0.71 with DINOv2 fixed, alongside planning gains. These results highlight the contribution of predictor supervision to action-effect prediction and planning.


\section{Conclusion}
\label{sec:conclusion}


JEPA world models are trained to predict logged transitions but used to compare actions, and behavior data leave this comparison unconstrained wherever the policy barely varies its action. Fork-JEPA measures the comparison directly, running two actions from the same restored state and matching the predicted difference to the measured one, with behavior prediction as an anchor and lookback to keep the state. With the same planner, its predictions carry the action effect at every rollout depth, and it plans better on three visual control tasks. However, each two-action fork provides only one action contrast per state, limiting local directional coverage. Future work could broaden coverage through multi-fork comparisons of three or more actions per state and active selection of fork states and directions.

\section*{LLM Usage Statement}
During the preparation of this work, the authors used OpenAI GPT-6.1 and Claude Opus-5.5 to assist with editing and organizing author-produced manuscript text, obtaining reviewer-style feedback, and refining existing figures and figure layouts. After using these tools, the authors reviewed and edited the content as needed and take full responsibility for the content of the publication.

{
    \small
    \bibliographystyle{splncs04}
    \bibliography{main}
}

\clearpage
\clearpage
\appendix
\setcounter{equation}{0}
\setcounter{figure}{0}
\setcounter{table}{0}
\setcounter{algorithm}{0}
\renewcommand{\theequation}{S\arabic{equation}}
\renewcommand{\thetable}{S\arabic{table}}
\renewcommand{\thefigure}{S\arabic{figure}}
\renewcommand{\thealgorithm}{S\arabic{algorithm}}
\maketitlesupplementary

\begin{center}
\textbf{SUMMARY OF THE APPENDIX}
\end{center}
\noindent This supplement provides proofs, implementation details, and additional experiments.
\begin{itemize}
\item \cref{sec:theory} presents \textbf{proofs and theory}, which include proofs of the three main-paper propositions and the fork budget behind the isotropic design.
\item \cref{sec:impl} gives \textbf{implementation details}: training, fork collection, restoration, evaluation, and pseudocode (\cref{sec:algo}).
\item \cref{sec:probes} describes the \textbf{action-effect probes}, which include the rollout probes, the position readouts on real frames, and the TwoRoom wall grid.
\item \cref{sec:plan} reports \textbf{planning results}, which include success rates, candidate ranking, and difference versus direct fork targets, and the baselines of the lever comparison.
\item \cref{sec:b9} presents the \textbf{regime world}, which includes the world and its evaluation, the predicted action response, the state maps of the component ablation, forks versus more behavior data, planning examples, forks on a sparse latent, and the sensitivity to the fork start.
\item \cref{sec:voe-supp} covers \textbf{violation of expectation}, with the surprise curves, the protocol, and the raw errors.
\item \cref{sec:more} presents \textbf{additional results}, which include the action effect along the rollout.
\item \cref{sec:todo} presents \textbf{additional controls and ablations}, covering data, training, mechanisms, alternatives, and cost.
\item \cref{sec:limits} discusses \textbf{limitations and future work}, including real-hardware deployment and multi-branch forks.
\end{itemize}

\section{Proofs and Theory}
\label{sec:theory}
This section proves main-paper Propositions~1--3, gives the supplementary probe-transfer bound, and adds the budget analysis behind the isotropic fork design. Throughout, $s$, $a$, and $y$ are a local state, action, and response, $\delta$ is a fork difference, and $G_\Delta=\sum_{i=1}^{P}\delta_i\delta_i^\top$ is the fork design matrix.

\subsection{Proof of Proposition 1}
\label{sec:a1}
Under the feedback model $a_t=-Mr(s_t)+\zeta_t$, project the behavior design vector $\phi_t=[r(s_t);a_t]$ onto $v_w=[M^\top w;w]$:
\begin{equation}
v_w^\top \phi_t
=w^\top Mr(s_t)+w^\top a_t=w^\top\zeta_t.
\label{eq:s1}
\end{equation}
Summing the squares over $t$ gives the identity of Proposition~1. With $\mathbb E[\zeta\mid s]=0$ and a fixed state distribution,
\begin{equation}
v_w^\top\mathbb E[G_N]v_w=Nw^\top\Sigma_aw.
\label{eq:s2}
\end{equation}
The identity needs no independence between samples; spectral bounds and parameter rates need the usual mixing and moment assumptions on top.

\noindent\textbf{Unexcited directions.} If $w\in\ker\Sigma_a$, the nonnegative variable $(w^\top\zeta_t)^2$ has zero mean, so it vanishes almost surely and $G_Nv_w=0$ for any finite sample. If $\Sigma_a=0$, this holds for every $w$, and since $w\mapsto v_w$ is injective, $\ker G_N$ has dimension at least $m$. Under deterministic linear feedback with state term $B_rr(s)$, the parameters $(B_r+\Xi M,\,B_a+\Xi)$ then give the same response on every behavior action. Partial excitation confines ambiguity to unexcited directions.

\noindent\textbf{No estimator closes the gap.} Suppose two environments induce the same distribution over behavior histories but respond differently to a candidate $e$, with responses $q_1(e)$ and $q_2(e)$. Any estimator $W$ computed from the histories has the same distribution in both, so
\begin{equation}
\|q_1(e)-q_2(e)\|
\le\mathbb E\|W-q_1(e)\|+\mathbb E\|W-q_2(e)\|,
\label{eq:s3}
\end{equation}
and one of the two expected errors is at least half the gap. Prediction, representation regularization, and lookback trained on behavior data alone cannot remove it. Forks add the missing same-state variation.

\subsection{Proof of Proposition 2}
\label{sec:a2}
At a shared state (A2), the state term cancels:
\begin{equation}
y(s,a)-y(s,a')=B_a(a-a')+\epsilon-\epsilon'.
\label{eq:s4}
\end{equation}
Write the predictor as $\hat y(a)=\hat u+Ba$. With $\delta=Uz$ and $C=\mathbb E[zz^\top]\succ0$ from (A3), the population fork-pair loss is
\begin{equation}
\begin{aligned}
    \mathcal L(B) &=\mathbb E\|B\delta-B_a\delta-\Delta\epsilon\|_2^2 \\
    &=\operatorname{tr}\!\big((B-B_a)U C U^\top(B-B_a)^\top\big)+\mathrm{const},
\end{aligned}
\label{eq:s4b}
\end{equation}
where the constant is $\mathbb E\|\Delta\epsilon\|^2$ and the cross term vanishes by (A1). Since $C\succ0$, the first term is zero exactly when $(B-B_a)U=0$. Every minimizer therefore satisfies $BU=B_aU$, and the orthogonal complement of $\mathrm{colspan}(U)$ stays free. Positive definiteness of $U^\top G_\Delta U$ alone would not be enough without the support condition in (A3). If $B_a$ varies with the state, the same argument applies locally wherever the forks cover the action space.

\noindent\textbf{What the label contains.} If the two restorations land at $s$ and $s'$, the difference also contains $u(s)-u(s')$, which the loss reads as an action effect; \cref{sec:restore-sens} measures how much this matters. Independent branch noise with covariance $\Sigma_\epsilon$ gives the difference a noise covariance of $2\Sigma_\epsilon$, which sets the estimation error (\cref{sec:a5}).

\subsection{Proof of Proposition 3}
\label{sec:a3}
In the deterministic, fully observed simplification, let $Q$ embed states into latents with $Q^\top Q=I$, so that the encoding errors are $\|E(o_0)-Qs_0\|,\|E(o_g)-Qs_g\|\le\varepsilon_{\rm rep}$. Write $\hat h_T=F^H(E(o_0),\mathbf a)$, $h_g=E(o_g)$, and $s_T=f^H(s_0,\mathbf a)$. The Lipschitz condition gives
\begin{equation}
\begin{aligned}
\|\hat h_T-Qs_T\|
&\le
\|F^H(E(o_0),\mathbf a)-F^H(Qs_0,\mathbf a)\|
+\varepsilon_{\rm tr}(\mathbf a)\\
&\le L^H\varepsilon_{\rm rep}+\varepsilon_{\rm tr}(\mathbf a).
\end{aligned}
\label{eq:s7}
\end{equation}
Let $u=\hat h_T-h_g$ and $v=Q(s_T-s_g)$, so that $\|u-v\|\le \beta(\mathbf a):=(1+L^H)\varepsilon_{\rm rep}+\varepsilon_{\rm tr}(\mathbf a)$. Since $Q^\top Q=I$ and $\|u\|,\|v\|\le R$,
\begin{equation}
|\hat c(\mathbf a)-c(\mathbf a)|
=|\langle u-v,u+v\rangle|
\le2R\,\beta(\mathbf a).
\label{eq:s8}
\end{equation}
Let $\mathbf a^\star\in\arg\min_{\mathcal C}c$. Since $\hat{\mathbf a}$ minimizes $\hat c$,
\begin{equation}
\begin{aligned}
c(\hat{\mathbf a})-c(\mathbf a^\star)
&\le|\hat c(\hat{\mathbf a})-c(\hat{\mathbf a})|
+|\hat c(\mathbf a^\star)-c(\mathbf a^\star)|\\
&\le4R\max_{\mathcal C}\beta(\mathbf a),
\end{aligned}
\label{eq:s9}
\end{equation}
which is Eq.~(6). If instead every terminal point and the goal lie in a ball of radius $R$, the difference vectors are bounded by $2R$ and the constant becomes $8R$.

\noindent\textbf{Why the anchor is needed.} An exact difference does not bound the regret. Let the true successors be $0$ and $1$, the goal $0$, and the predicted successors $-1$ and $0$. The predicted difference is exact, yet the preferred action flips. Exact behavior prediction fixes the logged successor and rules out this common shift; residual anchor error can remain after training (\mainref{Sec.}{sec:loss}{3.4}).

\subsection{Probe-Transfer Bound and Proof}
\label{sec:b4}
\label{sec:proof-probe}
\noindent\textbf{Supplementary result (Probe Transfer).} Let $g(h)=Wh+b$ fit measured features $h$ to standardized targets $\tilde x$, with corresponding predictions $\hat h$. Then
\begin{equation}
\sqrt{1-R^2_{\rm pred}}\le\sqrt{1-R^2_{\rm meas}}+\|W\|_2\sqrt{\mathbb E\|\hat h-h\|^2/V},
\label{eq:probe-supp}
\end{equation}
where $R^2_{\rm pred}=R^2(\tilde x,g(\hat h))$, $R^2_{\rm meas}=R^2(\tilde x,g(h))$, and $V=\mathbb E\|\tilde x-\mathbb E\tilde x\|^2$.
\noindent The bound separates the readability of measured features from the prediction mismatch, which fork supervision reduces for \MIDD{}.

The readout $g$ is fitted on measured features only, so it is fixed when applied to predictions. Since $g(\hat h)-g(h)=W(\hat h-h)$, the triangle inequality in $L^2$ gives
\begin{equation}
\begin{aligned}
\sqrt{\mathbb E\|\tilde x-g(\hat h)\|^2}
&\le\sqrt{\mathbb E\|\tilde x-g(h)\|^2}+\sqrt{\mathbb E\|W(\hat h-h)\|^2}\\
&\le\sqrt{\mathbb E\|\tilde x-g(h)\|^2}+\|W\|_2\sqrt{\mathbb E\|\hat h-h\|^2}.
\end{aligned}
\label{eq:s-probe}
\end{equation}
With standardized targets, $1-R^2(\tilde x,\hat x)=\mathbb E\|\tilde x-\hat x\|^2/V$ for the pooled $R^2$, and dividing by $\sqrt V$ gives \cref{eq:probe-supp}. For the uniformly averaged per-dimension $R^2$ used by \MID{} and \MIDD{}, apply the same argument after dividing each target coordinate and readout output by that coordinate\textquotesingle s test-set standard deviation; $W$ and $V$ must be expressed in those coordinates. Fitting-set standardization alone does not make test-set variances equal. \cref{sec:rollout-probes} evaluates both terms.

\subsection{Fork Budget and Design Energy}
\label{sec:a5}
\label{sec:design-energy}
\noindent\textbf{How much a budget can add.} Let $U$ be an orthonormal basis of a subspace $\mathcal S$ of dimension $m_{\rm eff}$, and let every executed difference satisfy $\|\delta_i\|\le\eta$. Then
\begin{equation}
\lambda_{\min}(U^\top G_\Delta U)
\le\frac{\operatorname{tr}(U^\top G_\Delta U)}{m_{\rm eff}}
=\frac{\sum_i\|U^\top\delta_i\|^2}{m_{\rm eff}}
\le\frac{P\eta^2}{m_{\rm eff}}.
\label{eq:s5}
\end{equation}
Equality holds when all differences lie in $\mathcal S$, have norm $\eta$, and form an isotropic tight frame. This is why the main paper draws $\delta$ isotropically over the whole block, so that each direction receives about $P/m$ of the budget listed in \mainref{Tab.}{tab:span}{1}. The rank of $U^\top G_\Delta U$ is at most $\min(P,m_{\rm eff})$, so a single pair covers one direction and fewer than $m_{\rm eff}$ pairs leave the smallest eigenvalue at zero. A full-rank marginal Gram matrix of behavior actions says nothing here, because it can have full rank without any variation at a fixed state.

\noindent\textbf{With an existing design.} If a design $G_0$ already exists with eigenvalues $\lambda_i$ on $\mathcal S$, the relaxed problem $\max_{n_i\ge0,\sum_i n_i=P}\min_i(\lambda_i+\eta^2n_i)$ uses water filling,
\begin{equation}
n_i=\frac{(\ell-\lambda_i)_+}{\eta^2},\qquad
\sum_i n_i=P,
\label{eq:s6}
\end{equation}
so the budget goes to the weakest directions first. Integer budgets and constrained action sets may not reach this optimum, which points to active fork selection as future work.

\noindent\textbf{Estimation error.} For a correctly specified linear model $Y=BX+E$ with a fixed full-row-rank design $X\in\mathbb R^{p\times N}$ and i.i.d.\ noise of variance $\sigma^2$, the least-squares error along a design vector $x$ is
\begin{equation}
\mathbb E[\|(\widehat B-B)x\|^2\mid X]
=d_y\sigma^2x^\top(XX^\top)^{-1}x
\le\frac{d_y\sigma^2\|x\|^2}{\lambda_{\min}(XX^\top)}.
\label{eq:s14}
\end{equation}
The error falls as $O(N^{-1})$ when $\lambda_{\min}$ grows linearly in $N$, which links \cref{eq:s5} to accuracy. For forks, $\sigma^2$ is the variance of the difference noise.

\begin{table}[t]
\supptabsetup
\setlength{\tabcolsep}{5pt}
\exptabular{5}{ccccc}{
\toprule
\rowcolor{tabHead}
\bfseries\boldmath $m$ & \bfseries\boldmath $P$ & \bfseries\boldmath E-optimal & \bfseries\boldmath Random & \bfseries\boldmath $P/m$\\
\midrule
2 & 128 & 128.00 & 67.17 & 64.00\\
3 & 128 & 128.00 & 40.42 & 42.67\\
4 & 128 & 128.00 & 33.21 & 32.00\\
5 & 128 & 128.00 & 25.79 & 25.60\\
5 & 32  & 32.00  & 6.03  & 6.40\\
\bottomrule
}
\caption{\textbf{Fork-design energy.} Median smallest projected eigenvalue over three seeds; $P/m$ is the isotropic reference.}
\label{tab:design}
\end{table}

\noindent\textbf{Controlled design setup.} \cref{tab:design} uses $d=64$, $m_{\rm plan}=1$, and $\eta=1$. The projected planning subspace in \cref{eq:s5} therefore has $m_{\rm eff}=1$, giving the bound $P$. E-optimal selection targets that direction; isotropic selection allocates $P/m$ across the $m$ action dimensions.

\noindent\textbf{Design comparison.} In the setup above, \cref{tab:design} illustrates \cref{eq:s5}: E-optimal selection concentrates the budget on the planning direction, while random isotropic selection distributes it across all action dimensions.

\subsection{Backward Pairing}
\label{sec:a4}
We also examined a paired version of lookback, in which both branches recover the shared predecessor $h_t$:
\begin{equation}
\mathcal L_{\rm lb}^{\rm pair}
=\|R(h_1^a,\tilde a)-h_t\|^2+
\|R(h_1^{a'},\tilde a')-h_t\|^2,
\label{eq:s10}
\end{equation}
with $\tilde a,\tilde a'$ the normalized actions the head receives. If $R(\cdot,\tilde a')$ is $L_R$-Lipschitz and $s_R=\|R(h_1^a,\tilde a)-R(h_1^a,\tilde a')\|$ measures how much the head uses its action, the triangle inequality gives $s_R\le\sqrt{2\mathcal L_{\rm lb}^{\rm pair}}+L_R\|h_1^a-h_1^{a'}\|$, hence
\begin{equation}
\|h_1^a-h_1^{a'}\|
\ge \frac{(s_R-\sqrt{2\mathcal L_{\rm lb}^{\rm pair}})_+}{L_R}.
\label{eq:s11}
\end{equation}
The bound separates the two successors only if the head responds to its action, and nothing in training forces it to, since a latent that stores the action lets the head ignore its action input. We therefore do not use this term; its weight is zero in every reported result.

\subsection{Symplectic Dynamics and Action Identification}
\label{sec:symplectic-identification}
The ambiguity in \cref{sec:a1} persists even with an identity encoder and an exactly symplectic predictor. Consider the separable Hamiltonian $H_b(q,p;a)=p^2/2-baq$, with scalar action $a$ held fixed over a step $\tau>0$. Its exact flow, also given exactly by kick--drift--kick leapfrog, is
\begin{equation*}
T_b(q,p;a)=\big(q+\tau p+\tfrac12\tau^2ba,\ p+\tau ba\big).
\end{equation*}
For every $b$, its state Jacobian is $A=\left(\begin{smallmatrix}1&\tau\\0&1\end{smallmatrix}\right)$, satisfying $A^\top JA=J$ for $J=\left(\begin{smallmatrix}0&1\\-1&0\end{smallmatrix}\right)$ and $\det A=1$. Its inverse is
\begin{equation*}
R_b(Q,P;a)=\big(Q-\tau P+\tfrac12\tau^2ba,\ P-\tau ba\big).
\end{equation*}
Thus $R_b(T_b(x;a);a)=x$ for every state and action, where $x=(q,p)$.

\noindent\textbf{Counterexample.} Let the true parameter be $b_\star$ and the behavior action always be zero. Every candidate $F=T_b$, $R=R_b$ then has zero behavior prediction and lookback error, since $T_b(x;0)=T_{b_\star}(x;0)$. Every candidate is also symplectic, invertible, and globally cycle-consistent. With $c_\tau=(\tau^2/2,\tau)^\top$, the action difference is
\begin{equation*}
T_b(x;a)-T_b(x;a')=c_\tau b(a-a'),
\end{equation*}
which disagrees with the truth whenever $b\ne b_\star$ and $a\ne a'$. Hence neither behavior lookback nor structural reversibility identifies an unexcited action response, even together. The ambiguity persists without excitation.

\noindent\textbf{Measured forks remove this ambiguity.} With the same identity encoder, the population fork loss in this family is
\begin{equation*}
\mathcal L_{\rm fork}(b)
=\|c_\tau\|^2(b-b_\star)^2\mathbb E[(a-a')^2].
\end{equation*}
If $\mathbb E[(a-a')^2]>0$, its unique minimizer is $b=b_\star$. This is an instance of the sampled-span result in \cref{sec:a2}. In the full model, lookback constrains latent predecessor recoverability, measured forks constrain action differences on their covered span, and behavior prediction anchors absolute successors (\mainref{Sec.}{sec:loss}{3.4}). These conclusions require the stated local-model and coverage assumptions; identifying physical consequences additionally requires a representation that retains the relevant physical differences.

\section{Implementation Details}
\label{sec:details}
\label{sec:impl}

\subsection{Model and Training}
\label{sec:b2}
LeWM and Fork-JEPA use a ViT-Tiny encoder trained from scratch on RGB frames, latent width 192, and a predictor conditioned on three frames and two preceding action blocks. The main paper abbreviates this input as $F(h_t,a_t)$. Actions are standardized per dimension and grouped into five-step blocks: $m=10$ on TwoRoom/PushT and $m=25$ on Cube. Training uses one NVIDIA H200 GPU.

Each update uses 128 behavior clips for prediction, SIGReg, and lookback, plus a separate fork batch for $\mathcal L_{\rm fork}$ (\mainref{Fig.}{fig:framework}{2}). Only the fork target $\Delta h$ is detached; behavior targets and both reverse-head inputs and targets retain gradients. Lookback recovers each predecessor from its successor and action. Fork supervision uses one-step pairs. LeWM and Fork-JEPA train from random initialization for 10 epochs with AdamW, weight decay $10^{-3}$, bf16, and SIGReg weight 0.09. \cref{tab:config} gives task settings; other baselines and frozen controls are described separately.

\begin{table}[t]
\supptabsetup
\setlength{\tabcolsep}{4.2pt}
\exptabular{7}{lcccccr}{
\toprule
\rowcolor{tabHead}
\bfseries\boldmath Task & \bfseries\boldmath LR & \bfseries\boldmath Clip & \bfseries\boldmath $\lambda_{\rm look}$ & \bfseries\boldmath $\lambda_{\rm fork}$ & \bfseries\boldmath Fork batch & \bfseries\boldmath $P$\\
\midrule
TwoRoom & $5\times10^{-5}$ & 1.0 & 1.0 & 1.0  & 32  & 1,000\\
PushT   & $2\times10^{-5}$ & 0.5 & 1.0 & 0.01 & 128 & 2,000\\
Cube    & $5\times10^{-5}$ & 0.5 & 1.0 & 0.01 & 128 & 1,000\\
Regime  & $5\times10^{-5}$ & 1.0 & 1.0 & 0.1  & 32  & 1,024\\
\bottomrule
}
\caption{\textbf{Task-specific training settings.} LR: learning rate; Clip: gradient-norm limit; Fork batch: pairs per update; $P$: training pairs.}
\label{tab:config}
\end{table}

\subsection{Fork Collection and Restoration}
\label{sec:b5}
\noindent\textbf{Collection.} Roots come from training episodes, excluding probe states and planning-evaluation episodes. At each root, restore the logged state and execute $a$, then restore again and execute $a'=a-\delta$, consistent with Proposition~2. Draw $\delta$ isotropically over the action block and clip the alternative to the action box. Record pre- and post-clipping actions; all losses and coverage statistics use executed values.

\noindent\textbf{What is restored.} Restore positions, velocities, object poses, goals, contacts, termination flags, and controller state. Omissions introduce $u(s)-u(s')$ into the label. On 200 Reacher roots, including the target that reset would resample, replay gives relative pixel error 0.0199 (threshold 0.02) and maximum joint-position/velocity error $4.45\times10^{-16}$.

\begin{table}[t]
\supptabsetup
\exptabular{2}{lc}{
\toprule
\rowcolor{tabHead}
\bfseries\boldmath Restoration check (Reacher) & \bfseries\boldmath Value\\
\midrule
Replayed roots & 200\\
Relative pixel error & 0.0199\\
Pixel-error threshold & 0.02\\
Maximum joint position/velocity error & $4.45\times10^{-16}$\\
\bottomrule
}
\caption{\textbf{Reacher restoration check.} Replay errors over 200 roots, including the maximum joint-state discrepancy.}
\label{tab:restore-check}
\end{table}

\begin{table}[t]
\supptabsetup
\setlength{\tabcolsep}{4.5pt}
\exptabular{7}{lcccccc}{
\toprule
\rowcolor{tabHead}
\bfseries\boldmath Task & \bfseries\boldmath $m$ & \bfseries\boldmath $P/m$ & \bfseries\boldmath Rank & \bfseries\boldmath Cond. & \bfseries\boldmath $\lambda_{\min}/\bar\lambda$ & \bfseries\boldmath Same outcome\\
\midrule
TwoRoom & 10 & 100 & 10 & 1.7 & 78\% & 1.2\%\\
PushT   & 10 & 200 & 10 & 1.4 & 84\% & 0.3\%\\
Cube    & 25 & 40  & 25 & 2.2 & 65\% & 0.8\%\\
\bottomrule
}
\caption{\textbf{Executed fork coverage.} Cond.: condition number; $\lambda_{\min}/\bar\lambda$: smallest-to-mean eigenvalue ratio; Same outcome: near-identical physical successors.}
\label{tab:coverage}
\end{table}

\noindent\textbf{Coverage definitions and scope.} Same outcome in \cref{tab:coverage} means a physical successor difference below 5\% of the median. Action rank does not establish state/contact coverage; the Reacher check validates only its own restoration.

\noindent\textbf{Coverage.} The executed differences span each task's action block, with $\lambda_{\min}/\bar\lambda\ge65\%$ and at most 1.2\% near-identical outcomes (\cref{tab:coverage}). Cube has the lowest budget per direction ($P/m=40$).

\subsection{Planning and Evaluation}
\label{sec:b3}
The architecture and the planner are those of LeWM, and only $E$ and $F$ run at test time.
CEM uses 300 candidates, 30 elites, variance coefficient 1, and 30 iterations on PushT or 10 on TwoRoom/Cube. Plans span five five-step blocks; execute one block before replanning, with goals 25 steps ahead. Visual-task SR uses 50 shared development episodes over three training seeds. World models share task-specific planner settings and evaluation seeds; goal-conditioned policies act directly on the same episodes. Training and probe fitting exclude planning-evaluation episodes. The reverse head is saved but unused at test time.

\subsection{Pseudocode}
\label{sec:algo}
\cref{alg:fork} collects fork futures and \cref{alg:train} runs one optimizer step of Fork-JEPA. Actions are in normalized units, and both use the executed actions after clipping. Planning is unchanged from LeWM and uses only $E$ and $F$.

\begin{algorithm*}[t]
\caption{Fork collection}
\label{alg:fork}
\begin{lstlisting}[style=pytorch]
# D: behavior data whose simulator states can be restored
# P: retained pairs (1,000 on TwoRoom and Cube, 2,000 on PushT)
# m: dimension of an action block; box: action box
# rho: size of the fork difference, in normalized action units

def collect_forks(D, P):
    forks = []
    # roots exclude the probe states and the planning-evaluation episodes
    for s in sample(D.roots, len(D.roots), replace=False):
        if len(forks) == P:
            break
        o_ctx, a_ctx = D.context(s)              # frames o_{t-2:t}, blocks a_{t-2:t-1}
        a = D.logged_block(s)                    # logged block a_t
        delta = rho * normalize(randn(m))        # isotropic over the whole block
        a_alt = clip(a - delta, box)             # executed alternative
        env.restore(s); o = env.step(a)          # one block of five steps
        env.restore(s); o_alt = env.step(a_alt)
        if not (env.done(o) or env.done(o_alt)):
            forks.append((o_ctx, a_ctx, a, a_alt, o, o_alt))
    return forks
\end{lstlisting}
\end{algorithm*}

\begin{algorithm*}[t]
\caption{One Fork-JEPA training step}
\label{alg:train}
\begin{lstlisting}[style=pytorch]
# E: encoder (ViT-Tiny and projector); F: causal predictor; R: reverse head R_psi
# B = 128 behavior clips; B_f: fork pairs per step (Tab. S2)
# lam_sig = 0.09; lam_look = 1.0; lam_fork: Tab. S2

def train_step(D, forks):
    #===== behavior batch =====#
    o, a = D.sample(B)                    # o: frames o_{t-2:t+1} (B x 4), a: blocks a_{t-2:t} (B x 3)
    h = E(o)                              # B x 4 x 192
    # L_base, Eq. (1): next latent at every position of the window; targets keep their gradient
    L_pred = mse(F(h[:, :3], a), h[:, 1:])
    L_sig = SIGReg(h)
    # L_look, Eq. (4): recover the predecessor from the successor and the action
    L_look = mse(R(h[:, 1:], a), h[:, :3]) \
           + mse(R(h[:, 2:], a[:, 1:]), h[:, 1:3])

    #===== fork batch =====#
    o_ctx, a_ctx, a, a_alt, o_next, o_alt = forks.sample(B_f)
    h_ctx = E(o_ctx)                      # root clip o_{t-2:t}, gradient kept
    dh_pred = F(h_ctx, cat(a_ctx, a))[:, -1] - F(h_ctx, cat(a_ctx, a_alt))[:, -1]
    dh = (E(o_next) - E(o_alt)).detach()  # measured difference, stop-gradient
    L_fork = mse(dh_pred, dh)             # Eq. (3)

    loss = L_pred + lam_sig * L_sig + lam_fork * L_fork + lam_look * L_look  # Eq. (5)
    optimizer.zero_grad()
    loss.backward()                       # E, F, R, and the action encoder
    clip_grad_norm_(params, clip)         # 1.0 or 0.5, Tab. S2
    optimizer.step()                      # AdamW, weight decay 1e-3
\end{lstlisting}
\end{algorithm*}

\section{Action-Effect Probes}
\label{sec:probes}
\label{sec:protocols}
We call a model \emph{transition-level identifiable} when a linear readout fitted on measured $\Delta h$ recovers physical changes $\Delta x$ and transfers to predicted $\Delta\hat h$.

Protocols and results extend \mainref{Sec.}{sec:effect}{4.2}.

\subsection{Rollout Probes}
\label{sec:rollout-probes}
\label{sec:b7}
\noindent\textbf{Protocol.} We evaluate frozen LeWM and Fork-JEPA checkpoints on 512 held-out roots. Each root forks at the first action block, and the two branches then share six continuation blocks, so $\Delta\hat h_k$ isolates how the first action's effect carries to depth $k$. Both branches of a root go to the same half of the data; one half fits the readouts and the other tests them. Features and targets are standardized on the fitting half. \MID{} fits the predicted successor latents to physical states, with a ridge penalty that leave-one-out cross-validation picks among eight values from $10^{-4}$ to $10^{3}$; this gives each model the best linear readout of its own predictions. \MIDD{} fits the measured branch differences, for which the intercept cancels, with RidgeCV over 25 values from $10^{-6}$ to $10^{6}$, and applies the frozen readout to the predicted differences. The physical target is the agent position on TwoRoom, the seven-dimensional state on PushT (agent and block positions, block angle, and agent velocity), and the full state observation of the robot and cube on Cube, which includes joint velocities and orientations; $R^2$ is averaged uniformly over the target dimensions. The $\delta$-only reference regresses $\Delta x_k$ on the executed first-block difference. Samples whose episode ends before depth $k$ are dropped at that depth. The main tables average depths 1 to 5, with $\pm$ over training seeds.

\begin{table}[t]
\supptabsetup
\setlength{\tabcolsep}{3.6pt}
\exptabular{7}{lcccccc}{
\toprule
\rowcolor{tabHead}
 & \multicolumn{2}{c}{\bfseries\boldmath measured $\Delta h$} & \multicolumn{2}{c}{\bfseries\boldmath predicted $\Delta\hat h$} & \multicolumn{2}{c}{\bfseries\boldmath deviation}\\
\cmidrule(lr){2-3}\cmidrule(lr){4-5}\cmidrule(l){6-7}
\rowcolor{tabHead}
\bfseries\boldmath Task & \bfseries\boldmath LeWM & \bfseries\boldmath Ours & \bfseries\boldmath LeWM & \bfseries\boldmath Ours & \bfseries\boldmath LeWM & \bfseries\boldmath Ours\\
\midrule
TwoRoom & 0.497 & \textbf{0.990} & $-0.049$ & \textbf{0.794} & -- & --\\
PushT   & 0.133 & \textbf{0.262} & 0.133 & \textbf{0.267} & 0.331 & 0.284\\
Cube    & 0.006 & \textbf{0.106} & 0.004 & \textbf{0.110} & 0.870 & 0.486\\
\bottomrule
}
\caption{\textbf{Measured and predicted action-effect readouts at depth 1.} Ours denotes Fork-JEPA; deviation is the normalized latent prediction mismatch.}
\label{tab:probedecomp}
\end{table}

\noindent\textbf{Prediction deviation.} Deviation in \cref{tab:probedecomp} is $\mathbb E\|\Delta\hat h-\Delta h\|^2/V$ in model-specific latent units. Deviation is reported for PushT and Cube.

\noindent\textbf{Where the effect is lost.} LeWM's TwoRoom measured score is 0.50 versus $-0.05$ on predictions, locating a predictor-side deficit. On PushT and Cube, both scores are low. Fork-JEPA improves measured and predicted readability (\cref{tab:probedecomp}).

\subsection{Position Readouts on Real Frames}
\label{sec:b8}
\label{sec:readout-frames}
For the encoder columns of \mainref{Tab.}{tab:probe}{2}, a linear readout and a two-layer MLP (256 and 256 ReLU units) map frozen encoder latents of real frames to object positions, with a 2,000/500/500 episode split; the validation episodes select the MLP's training epoch. \cref{tab:mlp-readout} reports the test error normalized by the label variance.

\begin{table}[t]
\supptabsetup
\setlength{\tabcolsep}{4pt}
\exptabular{8}{llcccccc}{
\toprule
\rowcolor{tabHead}
 &  & \multicolumn{3}{c}{\bfseries\boldmath linear $\downarrow$} & \multicolumn{2}{c}{\bfseries\boldmath MLP $\downarrow$} & \bfseries\boldmath Won\\
\cmidrule(lr){3-5}\cmidrule(lr){6-7}
\rowcolor{tabHead}
\bfseries\boldmath Task & \bfseries\boldmath Target & \bfseries\boldmath Rep. & \bfseries\boldmath LeWM & \bfseries\boldmath Ours & \bfseries\boldmath LeWM & \bfseries\boldmath Ours & \bfseries\boldmath (\%)\\
\midrule
TwoRoom & agent    & \rf{8}  & 8.2  & \textbf{0.1} & 2.7  & \textbf{0.2} & 97.8\\
\midrule
PushT   & agent    & \rf{52} & 44.1 & \textbf{7.2} & 14.6 & \textbf{6.7} & 89.0\\
        & block    & \rf{29} & \textbf{27.1} & 41.8 & \textbf{9.3} & 25.1 & 39.0\\
\midrule
Cube    & effector & \rf{18} & 21.2 & \textbf{3.5} & 12.6 & \textbf{5.2} & 90.4\\
        & block    & \rf{7}  & 8.1  & \textbf{7.0} & 5.1  & \textbf{4.2} & 56.0\\
\bottomrule
}
\caption{\textbf{Position readouts from measured frames.} Normalized test MSE ($\times10^3$; lower is better). Rep.: published LeWM error; Won: episodes (\%) with lower Fork-JEPA linear error.}
\label{tab:mlp-readout}
\end{table}

\noindent\textbf{The gain sits on the controlled body.} Fork-JEPA reduces linear error on the controlled body by 6--80 times and wins on at least 89\% of test episodes (\cref{tab:mlp-readout}); MLP readouts preserve this ordering. Passive-object gains are smaller: Cube-block error falls from 8.1 to 7.0, while PushT-block error rises from 27.1 to 41.8.

\begin{table}[t]
\supptabsetup
\setlength{\tabcolsep}{0.6pt}
\exptabular{6}{lccccc}{
\toprule
\rowcolor{tabHead}
\bfseries\boldmath \MIDD{} & \bfseries\boldmath $k=1$ & \bfseries\boldmath 2 & \bfseries\boldmath 3 & \bfseries\boldmath 4 & \bfseries\boldmath 5\\
\midrule
\rowcolor{gray!8}
\multicolumn{6}{@{}l}{\textbf{TwoRoom}}\\
\rf{$\delta$ only} & \rf{$0.93$} & \rf{$0.86$} & \rf{$0.79$} & \rf{$0.74$} & \rf{$0.70$}\\
LeWM & $-0.05$ & $-0.23$ & $0.00$ & $0.02$ & $-0.26$\\
Fork-JEPA & $\mathbf{0.79}$ & $\mathbf{0.78}$ & $\mathbf{0.69}$ & $\mathbf{0.62}$ & $\mathbf{0.58}$\\
\midrule
\rowcolor{gray!8}
\multicolumn{6}{@{}l}{\textbf{PushT}}\\
\rf{$\delta$ only} & \rf{$0.60$} & \rf{$0.30$} & \rf{$0.30$} & \rf{$0.28$} & \rf{$0.29$}\\
LeWM & $0.13$ & $0.07$ & $0.10$ & $0.06$ & $0.04$\\
Fork-JEPA & $\mathbf{0.27}$ & $\mathbf{0.30}$ & $\mathbf{0.31}$ & $\mathbf{0.33}$ & $\mathbf{0.31}$\\
\midrule
\rowcolor{gray!8}
\multicolumn{6}{@{}l}{\textbf{Cube}}\\
\rf{$\delta$ only} & \rf{$0.43$} & \rf{$0.20$} & \rf{$0.14$} & \rf{$0.13$} & \rf{$0.13$}\\
LeWM & $0.00$ & $0.01$ & $0.00$ & $0.00$ & $0.00$\\
Fork-JEPA & $\mathbf{0.11}$ & $\mathbf{0.13}$ & $\mathbf{0.05}$ & $\mathbf{0.02}$ & $\mathbf{0.02}$\\
\bottomrule
}
\caption{\textbf{Action-effect prediction by rollout depth.} \MIDD{} ($R^2$) at depths 1--5; gray rows show $\delta$-only scores.}
\label{tab:geometry}
\end{table}

\subsection{TwoRoom Wall Grid}
\label{sec:wall-grid}
\noindent\textbf{Grid and actions.} For the response maps of \mainref{Fig.}{fig:tworoom}{4}, we lay a $20\times20$ grid over the TwoRoom arena and keep the cells whose center lies in free space. At each kept cell $s$, we restore the agent at the center, at rest, with the task context of one fixed reference episode, and use three copies of the restored frame as history. The base block $a_0(s)$ is the logged block of the behavior state nearest to $s$, clipped step by step into the interior of the action box with a margin equal to the largest per-step change of the alternatives. The alternatives are $a_j=a_0+\eta\,\nu_j$ for $j=1,\dots,8$, where $\nu_j$ repeats the direction $(\cos\tfrac{2\pi j}{8},\sin\tfrac{2\pi j}{8})$ over the five steps and $\eta$ is half the median norm of the training fork differences. With this margin, every executed alternative equals its design, and all eight directions are kept in every cell.

\noindent\textbf{Responses and regions.} The true response is $\Delta x_j(s)=x(o^+_{a_j})-x(o^+_{a_0})$ and the predicted one $\Delta\hat x_j(s)=g^\Delta_1(F(c,a_j)-F(c,a_0))$, using each model's frozen depth-1 readout; no readout is fitted on the grid. A direction is blocked at $s$ if $\|\Delta x_j(s)\|$ is below 10\% of its median over open cells. A cell is open if no direction is blocked, a doorway cell if it lies in the door, and a wall cell otherwise. This gives 332 open, 25 wall, and 3 doorway cells; the figure reports the first two.

\noindent\textbf{Error.} The response error of a cell is $\mathrm{err}(s)=\sum_j\|\Delta\hat x_j(s)-\Delta x_j(s)\|^2/\sum_j\|\Delta x_j(s)\|^2$, summed over the eight directions, and \cref{tab:wall-grid} reports its median over cells. To separate the readout from the predictor, the measured error applies the same readout to the encoded measured successors, $g^\Delta_1(E(o^+_{a_j})-E(o^+_{a_0}))$. The $\delta$-only reference is a ridge map from the executed action difference to the physical response, fitted on the training roots of the depth-1 readout and never on the grid. \cref{tab:wall-grid}(a) reports the medians. LeWM's measured error is already 0.94, so most of its predicted error of 1.10 comes from a latent difference that the readout cannot map to the response; Fork-JEPA scores 0.021 and 0.029, respectively.

\noindent\textbf{State maps.} \mainref{Fig.}{fig:tworoom}{4} uses the released LeWM checkpoint and Fork-JEPA. Each 3D t-SNE map is viewed to minimize walk crossings and rotated onto the physical layout. These rigid transformations preserve distances. Among 360 free cells, 89\% of grid neighbors lie among the 8 nearest latent neighbors for LeWM, versus 100\% for Fork-JEPA. Cross-wall links account for 7.6\% and 0.2\% of their respective 8-neighbor graphs. LeWM preserves local neighborhoods but folds the rooms together.

\noindent\textbf{Where the readout is fitted.} We read the one-cell steps between neighboring free cells from the difference of their frozen latents at rest, with a linear map fitted ever farther from where it is used, namely a ridge map on every other step of the grid (nearby), a ridge map with the cell's $4\times4$ block left out (elsewhere on the grid), and the frozen depth-1 readout of the response maps (training data). A cell's error is its summed squared step error over its summed squared steps, as for $\mathrm{err}$. LeWM's median error grows from 0.24 to 0.48 to 1.09, while Fork-JEPA's stays at 0.011, 0.020, and 0.032. LeWM's latent thus holds the local displacement, but the map from latent difference to displacement changes across the arena, so a readout fitted on one part does not carry to another.

\section{Planning}
\label{sec:plan}
This section gives the numbers behind \mainref{Sec.}{sec:planning}{4.3} and two controls on how the planner uses the predictions.

\subsection{Success Rates}
\label{sec:sr-values}

\begin{table}[t]
\supptabsetup
\begin{minipage}[t]{\linewidth}
\vspace{0pt}
\centering\textbf{(a) TwoRoom wall-grid responses}\par\vspace{3pt}
\setlength{\tabcolsep}{4pt}
\exptabular{5}{llccc}{
\toprule
\rowcolor{tabHead}
\bfseries\boldmath Readout & \bfseries\boldmath Latents & \bfseries\boldmath all (360) & \bfseries\boldmath open (332) & \bfseries\boldmath wall (25)\\
\midrule
$\delta$ only & -- & 0.012 & 0.011 & 0.48\\
\midrule
\multirow{2}{*}{LeWM} & predicted & 1.10 & 1.11 & 0.99\\
 & measured & 0.94 & 0.97 & 0.93\\
\midrule
\multirow{2}{*}{Fork-JEPA} & predicted & 0.029 & 0.026 & 0.14\\
 & measured & 0.021 & 0.020 & 0.061\\
\bottomrule
}
\end{minipage}
\par\vspace{8pt}
\begin{minipage}[t]{\linewidth}
\vspace{0pt}
\centering\textbf{(b) Visual-task planning success}\par\vspace{3pt}
\setlength{\tabcolsep}{4pt}
\exptabular{4}{lccc}{
\toprule
\rowcolor{tabHead}
\bfseries\boldmath Method & \bfseries\boldmath TwoRoom & \bfseries\boldmath PushT & \bfseries\boldmath Cube\\
\midrule
\rowcolor{gray!8}
\multicolumn{4}{l}{\emph{Goal-conditioned policies}}\\
GCBC    & $100\pm0$ & $74\pm4$ & $78\pm2$\\
GCIQL   & $100\pm0$ & $24\pm8$ & $62\pm2$\\
GCIVL   & $100\pm0$ & $40\pm6$ & $52\pm2$\\
\rowcolor{gray!8}
\multicolumn{4}{l}{\emph{World models, same planner}}\\
PLDM    & $95\pm1$  & $74\pm8$ & $66\pm4$\\
DINO-WM & $\mathbf{100\pm0}$ & $75\pm6$ & $\mathbf{82\pm8}$\\
LeWM    & $86\pm2$  & $92\pm4$ & $66\pm4$\\
\hspace{.5em}+lookback & $\mathbf{100\pm0}$ & $82\pm8$ & $72\pm6$\\
\hspace{.5em}+forks    & $88\pm2$  & $\mathbf{94\pm2}$ & $76\pm2$\\
Fork-JEPA & $\mathbf{100\pm0}$ & $\mathbf{94\pm4}$ & $80\pm4$\\
\bottomrule
}
\end{minipage}
\caption{\textbf{TwoRoom responses and visual-task planning.} (a) Median relative response error (lower is better). (b) SR (\%), mean $\pm$ standard deviation over three seeds on 50 development episodes; bold marks the highest world-model SR.}
\label{tab:wall-grid}
\label{tab:sr-values}
\end{table}

\cref{tab:sr-values}(b) gives the values of \mainref{Fig.}{fig:sr}{5}. Fork-JEPA matches or exceeds the other world models trained from scratch on every task. On TwoRoom, lookback and Fork-JEPA reach 100\%, versus LeWM's 86\% (\chg{$+14$~pp}{86\%}{100\%}). On PushT, policy SR spans 24--74\%, LeWM reaches 92\%, and single-term gains remain within seed variation. On Cube, both terms improve LeWM and their combination is best among models trained from scratch; frozen DINO-WM reaches 82\%, comparable within seed spread.

\subsection{Candidate Ranking}
\label{sec:ranking}
\label{sec:pairwise}
\noindent\textbf{Protocol.} For each task, 200 roots outside the development episodes each receive 64 action sequences drawn from the initial CEM distribution, with the goal 25 steps later. We execute every sequence and measure its true endpoint cost after one and five blocks, both in the model's latent space and as physical distance to the goal. After restoration screening, the evaluation retains 194 TwoRoom roots, 200 PushT roots, and 197 Cube roots.

\begin{table}[t]
\supptabsetup
\exptabular{2}{ll}{
\toprule
\rowcolor{tabHead}
\bfseries\boldmath Candidate-ranking protocol & \bfseries\boldmath Setting\\
\midrule
Requested roots per task & 200\\
Executed candidates per root & 64\\
Candidate distribution & Initial CEM distribution\\
Goal offset & 25 environment steps\\
Scored depths & 1 and 5 action blocks\\
Root exclusions & 6 TwoRoom; 3 Cube\\
True-cost spaces & Latent and physical\\
Top-candidate summary & Top 6 at depth 1\\
\bottomrule
}
\caption{\textbf{Candidate-ranking evaluation settings.} Numerical results are in \cref{tab:ranking}.}
\label{tab:ranking-protocol}
\end{table}

\begin{table}[t]
\supptabsetup
\setlength{\tabcolsep}{2.8pt}
\renewcommand{\arraystretch}{1.12}
\exptabular{7}{lcccccc}{
\toprule
\rowcolor{tabHead}
 & \multicolumn{4}{c}{\bfseries\boldmath Kendall's $\tau_b$ $\uparrow$} & \multicolumn{2}{c}{\bfseries\boldmath top-6}\\
\cmidrule(lr){2-5}\cmidrule(l){6-7}
\rowcolor{tabHead}
 & \multicolumn{2}{c}{\bfseries\boldmath latent} & \multicolumn{2}{c}{\bfseries\boldmath physical} &  & \\
\cmidrule(lr){2-3}\cmidrule(lr){4-5}
\rowcolor{tabHead}
\bfseries\boldmath Model & \bfseries\boldmath $k=1$ & \bfseries\boldmath 5 & \bfseries\boldmath $k=1$ & \bfseries\boldmath 5 & \bfseries\boldmath \multirow{-2}{*}{overlap $\uparrow$} & \bfseries\boldmath \multirow{-2}{*}{regret $\downarrow$}\\
\midrule
\rowcolor{gray!8}
\multicolumn{7}{@{}l}{\emph{TwoRoom}}\\
LeWM      & 0.674 & 0.632 & 0.231 & 0.263 & 0.70 & 7.44\\
Fork-JEPA & \textbf{0.939} & \textbf{0.895} & \textbf{0.501} & \textbf{0.589} & \textbf{0.91} & \textbf{4.95}\\
\rowcolor{gray!8}
\multicolumn{7}{@{}l}{\emph{PushT}}\\
LeWM      & 0.777 & 0.749 & 0.157 & 0.241 & 0.75 & 49.9\\
Fork-JEPA & \textbf{0.821} & \textbf{0.786} & \textbf{0.235} & \textbf{0.378} & \textbf{0.77} & \textbf{44.9}\\
\rowcolor{gray!8}
\multicolumn{7}{@{}l}{\emph{Cube}}\\
LeWM      & 0.479 & 0.352 & \textbf{0.113} & \textbf{0.078} & 0.43 & \textbf{0.0279}\\
Fork-JEPA & \textbf{0.628} & \textbf{0.519} & 0.075 & 0.051 & \textbf{0.61} & 0.0280\\
\bottomrule
}
\caption{\textbf{Candidate ranking.} Mean Kendall\textquotesingle s $\tau_b$ at depths 1 and 5. Top-6 overlap is a fraction; regret is excess physical cost in task units. Higher correlation/overlap and lower regret are better.}
\label{tab:ranking}
\end{table}

\noindent\textbf{Ranking summaries.} We retain 194/200/197 roots on TwoRoom/PushT/Cube, each with 64 candidates. Top-6 overlap compares predicted and measured selections; regret is the depth-1 excess physical cost.

\noindent\textbf{Latent and physical candidate rankings.} Fork-JEPA improves mean latent ranking on all tasks at both depths (\cref{tab:ranking}). Physical ranking improves on TwoRoom/PushT but falls slightly on Cube, whose regret is nearly unchanged (0.0279 versus 0.0280). Thus latent ranking gains need not transfer to physical cost. \cref{tab:pairwise} tests two-branch choices.

\cref{sec:limits} explains why difference supervision alone does not guarantee correct candidate rankings.

\subsection{Difference versus Direct Fork Targets}
\label{sec:direct-vs-diff}

\begin{table}[t]
\supptabsetup
\setlength{\tabcolsep}{5pt}
\exptabular{7}{lcccccc}{
\toprule
\rowcolor{tabHead}
 & \multicolumn{2}{c}{\bfseries\boldmath TwoRoom} & \multicolumn{2}{c}{\bfseries\boldmath PushT} & \multicolumn{2}{c}{\bfseries\boldmath Cube}\\
\cmidrule(lr){2-3}\cmidrule(lr){4-5}\cmidrule(l){6-7}
\rowcolor{tabHead}
\bfseries\boldmath Target & \bfseries\boldmath SR & \bfseries\boldmath \MIDD{} & \bfseries\boldmath SR & \bfseries\boldmath \MIDD{} & \bfseries\boldmath SR & \bfseries\boldmath \MIDD{}\\
\midrule
direct & 88 & 0.576 & 80 & 0.274 & 72 & $-0.06$\\
diff   & \textbf{100} & \textbf{0.694} & \textbf{94} & \textbf{0.302} & \textbf{80} & \textbf{0.064}\\
\bottomrule
}
\caption{\textbf{Direct versus difference fork targets.} SR (\%) and depth-1--5 \MIDD{} with the same measured forks and training budget; higher is better.}
\label{tab:direct-vs-diff}
\end{table}

\noindent\textbf{The difference target matters on the visual tasks.} Difference matching improves SR by 8--14 percentage points over direct regression and raises \MIDD{} on all three tasks (\cref{tab:direct-vs-diff}). This is consistent with canceling shared state error. With lookback in the regime world, SR is similar (68\% versus 67\%), so the benefit depends on the setting.

\subsection{Baselines of the Lever Comparison}
\label{sec:lever-baselines}
For methods with +forks rows, \mainref{Tab.}{tab:paradigms}{4} adds measured fork supervision to the baseline objective while retaining its backbone, data, budget, and planner (\cref{sec:impl}). Fork pairs follow \mainref{Tab.}{tab:span}{1}, with $\lambda_{\rm fork}$ from \cref{tab:config}; lookback is not added. \cref{tab:imagined-forks,tab:proxy-baselines} describe imagined forks and isolated proxies.
\noindent\textbf{What differs between baseline families.} PLDM/LpWM change the objective; GA-WM adds privileged state labels; AdaJEPA adapts at test time. PhyLatent scores are externally reported, as marked in Main Table~4. The isolated proxies in \cref{tab:proxy-baselines} are not full GA-WM runs.
\noindent\textbf{Local implementation settings.} The settings for PLDM, LpWM, and GA-WM are detailed below; AdaJEPA settings follow its official implementation.

\noindent\textbf{PLDM.} Prediction weight is 1; SIGReg is replaced by batch standard-deviation, covariance, temporal standard-deviation, smoothness, and inverse-dynamics losses with base weights 3.9843, 6.9238, 0.24535, 0.74242, and 1.072. A shared multiplier is selected from $\{0.1,1,10\}$ on dev50. The local model uses four-frame clips with three-frame context and normalized five-step action blocks.

\noindent\textbf{LpWM.} RDMReg replaces SIGReg, using a ReLU link, sliced-Wasserstein loss, 1,024 slices, $p=1$, $\mu=0$, and target RMS normalization. Base weight 25 is scaled by $\{0.1,1,10\}$. Latents remain sparse; TJ is tested separately.

\noindent\textbf{Temporal Jaccard (TJ).} The local LpWM+TJ implementation adds $\lambda_{\rm TJ}\mathcal L_{\rm TJ}$ to the LpWM objective, where
\[
\mathcal L_{\rm TJ}=\left\langle 1-
\frac{\sum_j\min(h_{t,j},h_{t+1,j})}
{\sum_j\max(h_{t,j},h_{t+1,j})+10^{-8}}
\right\rangle_{b,t}.
\]
Here $h$ is the nonnegative sparse latent, and the average spans batch elements and adjacent frames. The registered control uses $\lambda_{\rm TJ}=0$ or $0.01$; positive weight encourages temporal support stability.

\noindent\textbf{GA-WM.} The local implementation retains SIGReg weight 0.09 and adds state alignment and inverse dynamics, each with base weight 1 and a shared dev50 multiplier in $\{0.1,1,10\}$. State targets have 2 channels on TwoRoom (position), 6 on PushT (four state coordinates and sine/cosine of block angle), and 33 on Cube (13 configuration, 2 gripper, 6 rotation, 6 arm-velocity, and 6 cube-velocity channels). The regime-world adaptation uses XY only, without zone labels. State-target statistics use behavior data, sample standard deviation, and a $10^{-6}$ floor. SA-only and IDM-only activate only their respective auxiliary loss.

\noindent\textbf{AdaJEPA.} The official protocol~\cite{wang2026adajepa} takes one Adam step between action execution and replanning, using the five latest executed transitions. Update the predictor's last Transformer block and final LayerNorm at $5\times10^{-4}$ and the encoder projection head at $10^{-5}$; freeze other parameters. Use mean one-step latent prediction loss with detached targets, history 3, and frameskip 5. Reset weights and buffer per episode.

\section{Dynamics-Regime World}
\label{sec:b9}
\label{sec:results}
This section gives the protocol and controls behind \mainref{Sec.}{sec:regimes}{4.4} and \mainref{Sec.}{sec:ablations}{4.6}.

\subsection{World and Evaluation}
\label{sec:regime-world}

\begin{figure}[t]
\centering
\includegraphics[width=\linewidth]{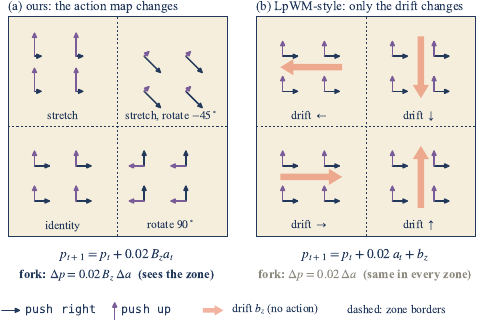}
\caption{\textbf{Zone dynamics.} (a) Our action-dependent regimes; (b) LpWM-style Piecewise dynamics. Rightward and upward pushes show the columns of $B_z$ with drift removed.}
\label{fig:regimefields}
\end{figure}

\noindent\textbf{Why we change the world.} Our four zones have distinct action matrices $B_z$ (\mainref{Fig.}{fig:regimemaps}{6}), no drift, and observations showing only the agent (\mainref{Sec.}{sec:regimes}{4.4}). LpWM's original Piecewise world instead shares one action matrix and varies drift $b_z$. In a same-state fork, drift cancels: original-world differences are zone-independent, whereas ours are $B_z\Delta a$ (\cref{fig:regimefields}). The adaptation therefore tests state-dependent action responses rather than offsets.

\noindent\textbf{Evaluation.} We plan from 200 fixed starts, 100 with a goal in the same zone and 100 with a goal across a zone boundary, and replan after every block. An episode succeeds when the agent ends within 0.05 of the goal.

\subsection{Predicted Action Response}
\label{sec:response}
\noindent\textbf{Protocol.} From each of 500 held-out positions, with the agent at rest, we apply 13 actions for one block each, namely pushes of $\pm0.25$ along each axis, eight more directions of the same size, and zero. We encode the true outcomes and predict them with the frozen predictor. An RBF kernel ridge readout maps measured latents to positions; within each zone the roots are split 60/20/20 for fitting, choosing the bandwidth and penalty, and testing. All readouts reach a position $R^2$ of at least 0.93 on validation. The two pairs of opposite pushes give an implied action map $\hat B$, from predicted or from measured outcomes, scored as $\|\hat B-B_z\|_F^2/\|B_z\|_F^2$ on the 82 test roots whose pushes stay inside one zone. We report medians to limit the influence of large errors.

\begin{table}[t]
\supptabsetup
\setlength{\tabcolsep}{3.2pt}
\textbf{(a) Action-map readout}\par\smallskip
\exptabular{7}{llccccc}{
\toprule
\rowcolor{tabHead}
 &  & \multicolumn{4}{c}{\bfseries\boldmath zone} & \\
\cmidrule(lr){3-6}
\rowcolor{tabHead}
\bfseries\boldmath Outcomes & \bfseries\boldmath Model & \bfseries\boldmath 1 & \bfseries\boldmath 2 & \bfseries\boldmath 3 & \bfseries\boldmath 4 & \bfseries\boldmath all\\
\midrule
\multirow{4}{*}{predicted} & LeWM      & 0.739 & \textbf{0.157} & \textbf{0.033} & 0.640 & 0.142\\
 & +lookback & 0.900 & 3.868 & 0.373 & 3.072 & 0.890\\
 & LpWM      & 0.186 & 0.261 & 0.084 & 0.257 & 0.189\\
 & Fork-JEPA & \textbf{0.153} & 0.221 & 0.035 & \textbf{0.109} & \textbf{0.087}\\
\midrule
\multirow{4}{*}{measured} & LeWM      & 0.118 & 0.192 & 0.024 & 0.030 & 0.043\\
 & +lookback & 0.075 & 0.153 & 0.053 & 0.058 & 0.069\\
 & LpWM      & 0.043 & 0.136 & 0.042 & 0.187 & 0.087\\
 & Fork-JEPA & 0.035 & 0.087 & 0.038 & 0.191 & 0.066\\
\bottomrule
}
\par\medskip
\textbf{(b) Historical sampling diagnostic}\par\smallskip
\exptabular{3}{lcc}{
\toprule
\rowcolor{tabHead}
\bfseries Sampling & \bfseries\shortstack{Cross-zone\\SR (\%) $\uparrow$} & \bfseries\shortstack{Worst-zone\\map error $\downarrow$}\\
\midrule
Independent transitions & 69 & 0.860\\
Same-state fork pairs   & 68 & 0.282\\
\bottomrule
}
\caption{\textbf{Regime-world action response and sampling diagnostic.} (a) Median normalized map error; all pools 82 test roots. (b) Historical checkpoints: cross-zone SR at $H=10$ on 100 shared starts, and worst-zone median map error using cached v4 readouts.}
\label{tab:response}
\label{tab:paired-independent}
\end{table}

\noindent\textbf{Aggregation.} In \cref{tab:response}(a), all pools 82 valid roots. Main Table~4 instead reports the worst zone median. The same measured-outcome readout evaluates predictions.

\noindent\textbf{Measured and predicted action maps.} Overall measured-outcome errors are low for all models (0.04--0.09). Fork-JEPA has the lowest overall predicted error (0.087 versus 0.142 for LeWM and 0.189 for LpWM) and the lowest error in zones 1 and 4; LeWM is best in zones 2 and 3. Lookback improves position readability but raises overall response error to 0.890, particularly in rotated zones. Regime recognition and action prediction are thus distinct (\cref{tab:response}).

\noindent\textbf{Sampling diagnostic and expectation.} \cref{tab:paired-independent}(b) reuses historical checkpoints with the same evaluation starts and planner. The observed SRs are similar; raw map error is lower for same-state forks. This diagnostic is descriptive and does not isolate the effect of sampling.

\subsection{State Maps of the Component Ablation}
\label{sec:ablation-maps}
\noindent\textbf{State-map construction.} All maps use the same $20\times20$ grid. Patches span four corner latents and fade between 1.5 and 4 times the median edge length. Colors encode position (gold/teal/navy/coral clockwise from upper left). The walk crosses lower-left, upper-left, and upper-right zones. Isomap uses eight neighbors without alignment; t-SNE is aligned to true position with a common row frame.

\begin{figure*}[t]
\centering
\includegraphics{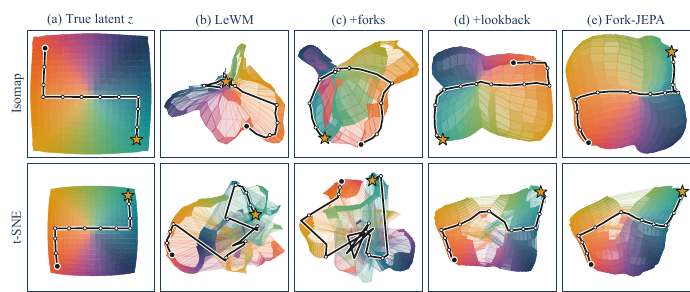}
\caption{\textbf{Component-ablation state maps.} Isomap (top) and t-SNE (bottom) on the same $20\times20$ state grid. Colors indicate position; translucent patches mark stretching. The walk crosses three zones from \legWalkStart{} to \legWalkGoal{}. Panel (a) uses true positions.}
\vspace{-4pt}
\label{fig:ablation-maps}
\end{figure*}

\cref{fig:ablation-maps} shows the arms of \mainref{Tab.}{tab:regimes}{3} on one $20\times20$ grid with a shared walk across three zones. Lookback produces a continuous sheet and smooth walk; LeWM and +forks remain folded. Adding forks to lookback improves the action response, reducing error by \chg{$-90\%$}{0.890}{0.087} (\cref{tab:response}). Fork-JEPA improves both properties.

\subsection{Forks versus More Behavior Data}
\label{sec:data-eff-values}

\begin{table}[t]
\supptabsetup
\setlength{\tabcolsep}{5pt}
\exptabular{5}{ccccc}{
\toprule
\rowcolor{tabHead}
\multicolumn{2}{c}{\bfseries\boldmath dense (SIGReg)} & \multicolumn{3}{c}{\bfseries\boldmath sparse (RDMReg)}\\
\cmidrule(lr){1-2}\cmidrule(l){3-5}
\rowcolor{tabHead}
\bfseries\boldmath LeWM & \bfseries\boldmath Full & \bfseries\boldmath LpWM & \bfseries\boldmath +lookback & \bfseries\boldmath Full\\
\midrule
35 & \textbf{68} & 41 & 48 & 64\\
\bottomrule
}
\caption{\textbf{Dense and sparse latent ablations.} Cross-zone SR (\%) at $H=10$. Full adds lookback and forks to the corresponding SIGReg or RDMReg baseline.}
\label{tab:ablate-grid}
\end{table}

\begin{table}[t]
\supptabsetup
\setlength{\tabcolsep}{5pt}
\exptabular{5}{lcccc}{
\toprule
\rowcolor{tabHead}
\bfseries\boldmath Behavior steps & \bfseries\boldmath 2.0M & \bfseries\boldmath 2.4M & \bfseries\boldmath 2.8M & \bfseries\boldmath 3.2M\\
\midrule
LeWM       & $35\pm2$ & $38\pm2$ & $40\pm8$ & $38\pm4$\\
+forks    & $48\pm2$ & $50\pm2$ & $54\pm8$ & $54\pm2$\\
+lookback & $61\pm4$ & $60\pm4$ & $62\pm6$ & $58\pm8$\\
Fork-JEPA  & $\mathbf{68\pm4}$ & $\mathbf{68\pm2}$ & $\mathbf{72\pm8}$ & $\mathbf{70\pm2}$\\
\bottomrule
}
\caption{\textbf{Behavior-data scaling.} Cross-zone SR (\%) at $H=10$, mean and spread over three seeds; fork models use 1,024 pairs.}
\label{tab:data-eff-values}
\end{table}

\noindent\textbf{Forks keep their gain at every data size.} Across 2.0M--3.2M behavior steps, LeWM changes by at most 5 SR points. Adding 1,024 forks gains 12--16 points over LeWM and 7--12 over lookback (\cref{tab:data-eff-values}). More behavior data may expand state coverage while leaving action directions unexcited.

\subsection{Planning Examples}
\label{sec:regime-examples}

\begin{figure}[t]
\centering
\includegraphics[width=\linewidth]{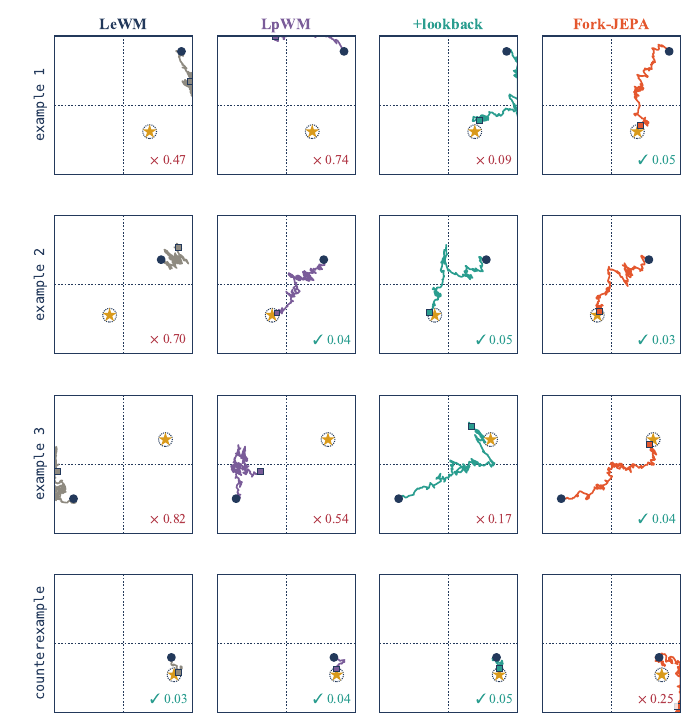}
\caption{\textbf{Regime-world planning examples.} Start \legStart{}, goal \legGoal{}, success radius \legRadius{} (0.05); dashed lines mark zone boundaries. Numbers give final goal distance.}
\label{fig:trajectories}
\end{figure}

\cref{fig:trajectories} shows three cross-zone examples where Fork-JEPA succeeds and LeWM fails, and a within-zone counterexample where LeWM succeeds but Fork-JEPA fails. These examples illustrate individual executions; aggregate planning results are reported in main Table~3.

\begin{figure}[t]
\centering
\includegraphics[width=\linewidth]{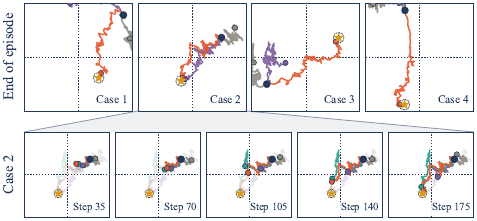}
\caption{\textbf{Executed trajectories across dynamics regimes.} Top: four cross-zone episodes at each model's termination. Bottom: Case 2 at control steps 35--175; faint lines show the complete paths. LeWM \legLine{figMuted}, LpWM \legLine{figPurple}, Fork-JEPA \legLine{figCoral}; the bottom row also includes +lookback \legLine{figTeal}. Start \legStart{}, goal \legGoal{}, success radius \legRadius{} (0.05).}
\label{fig:regime-trajectory-progress}
\end{figure}

\noindent\textbf{Progress across regime boundaries.} In cases 1, 3, and 4 of \cref{fig:regime-trajectory-progress}, Fork-JEPA reaches the goal while LeWM and LpWM terminate outside the success radius. Case 2 distinguishes eventual success from execution time: Fork-JEPA succeeds at step 175, lookback at 185, and LpWM at 225, while LeWM fails within the 250-step budget. The snapshots show Fork-JEPA progressing through the boundary region toward the goal while LeWM remains near its starting zone. These executed trajectories illustrate closed-loop behavior; they do not establish a causal attribution to a particular loss. The counterexample in \cref{fig:trajectories} also shows that the advantage is not universal.

\subsection{Forks on a Sparse Latent}
\label{sec:sparse}

\noindent\textbf{The fork signal does not depend on SIGReg.} Adding forks to sparse-latent lookback raises cross-zone SR from 48\% to 64\%, versus 68\% for the dense full model (\cref{tab:ablate-grid}). Forks complement both regularizers.

\subsection{Sensitivity to the Fork Start}
\label{sec:restore-sens}
\noindent\textbf{Protocol.} In the regime world the true dynamics are known, so we can move the two starts of a fork apart on purpose. In the first setting, Gaussian noise of standard deviation $\sigma$ is added to the physical start of the alternative branch only, with the logged branch, roots, actions, behavior data, and initialization fixed. In the second, there is no restoration, and a fork pairs two logged transitions whose starts lie within radius $r$. Both $\sigma$ and $r$ are in units of the median one-block displacement, 0.0545. We also report the offset, the median distance between the two starts divided by the median action effect $0.02\,\|B_z\textstyle\sum\delta\|$.

\begin{table*}[t]
\supptabsetup
\setlength{\tabcolsep}{5pt}
\renewcommand{\arraystretch}{1.12}
\exptabular{10}{cccccccccc}{
\toprule
\rowcolor{tabHead}
\multicolumn{2}{c}{\bfseries\boldmath fork starts} & \multicolumn{4}{c}{\bfseries\boldmath cross-zone SR (\%) $\uparrow$} & \bfseries\boldmath all SR (\%) $\uparrow$ & \multicolumn{2}{c}{\bfseries\boldmath state readout $\uparrow$} & \bfseries\boldmath map error $\downarrow$\\
\cmidrule(lr){1-2}\cmidrule(lr){3-6}\cmidrule(lr){8-9}
\rowcolor{tabHead}
\bfseries\boldmath $\sigma$ or $r$ & \bfseries\boldmath offset & \bfseries\boldmath $H=5$ & \bfseries\boldmath $10$ & \bfseries\boldmath $15$ & \bfseries\boldmath $20$ & \bfseries\boldmath $H=10$ & \bfseries\boldmath pos.\ $R^2$ & \bfseries\boldmath zone BA (\%) & \bfseries\boldmath worst zone\\
\midrule
\rowcolor{gray!8}
\multicolumn{10}{@{}l}{\emph{Restored, with noise $\sigma$ on the alternative branch}}\\
0    & 0    & 61 & 68 & 52 & 37 & 78.5 & 0.92 & 88.4 & 0.22\\
0.05 & 0.43 & 70 & 67 & 53 & 39 & 82.5 & 0.93 & 86.4 & 0.08\\
0.10 & 0.87 & 59 & 66 & 50 & 32 & 79.0 & 0.85 & 93.2 & 0.13\\
0.25 & 2.17 & 64 & 65 & 57 & 38 & 79.0 & 0.77 & 92.8 & 4.34\\
0.50 & 4.35 & 61 & 66 & 54 & 42 & 78.5 & 0.75 & 85.8 & 2.62\\
1.00 & 8.69 & 61 & 61 & 49 & 36 & 74.0 & 0.91 & 87.8 & 6.66\\
\midrule
\rowcolor{gray!8}
\multicolumn{10}{@{}l}{\emph{Logged neighbors within radius $r$, no restoration}}\\
0.05 & 0.26 & 61 & 62 & 47 & 38 & 76.5 & 0.88 & 92.0 & 0.31\\
0.10 & 0.51 & 59 & 59 & 50 & 34 & 78.0 & 0.89 & 92.2 & 0.87\\
0.25 & 1.26 & 65 & 65 & 56 & 37 & 78.0 & 0.88 & 86.4 & 0.17\\
0.50 & 2.50 & 62 & 66 & 49 & 41 & 75.5 & 0.89 & 87.8 & 0.66\\
\bottomrule
}
\caption{\textbf{Fork-start mismatch in the regime world.} SR and zone BA are percentages. Map error is the worst-zone median; lower is better. $\sigma$, $r$, and offset are defined in \cref{sec:restore-sens}.}
\label{tab:restore-sens-regime}
\label{tab:neighbor-forks}
\end{table*}

\noindent\textbf{Planning tolerates large offsets; the response map is more sensitive.} At $H=10$, cross-zone SR stays within 7 points of exact restoration up to offset 8.7, but worst-zone map error rises to 6.66 (\cref{tab:restore-sens-regime}). Logged-neighbor forks reach 59--66\% SR at $H=10$ without restoration. State mismatch can therefore harm response accuracy before planning success. Visual-task sensitivity is tested in \cref{sec:todo-data}.

\section{Violation of Expectation}
\label{sec:voe-supp}
\begin{figure}[t]
\centering
\includegraphics[width=\linewidth]{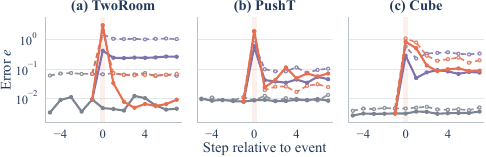}
\caption{\textbf{Surprise over time.} (a) TwoRoom, (b) PushT, (c) Cube. Normal \legLine{surpriseGray}, recolor \legLine{surprisePurple}, teleport \legLine{surpriseOrange}; Fork-JEPA \legSolidDot{}, LeWM \legDashedOpen{}. Events occur at step 0. Error scales are model-specific.}
\label{fig:surprise}
\end{figure}
A predictor that knows what actions can do should also notice what they cannot. We test this with LeWM's violation-of-expectation framework~\cite{leworldmodel} on 100 matched events per task that share the starting context and actions. A teleport of the controlled object is a change no action can cause, while a recolor changes only appearance~\cite{riochet2022intphys, garrido2025intuitive}.

\cref{fig:surprise} shows that both models are more surprised by teleport than by recolor on every task. The gap is largest on TwoRoom, where Fork-JEPA reads action effects best, and its teleport peak stands far above its recolor response, while LeWM responds to both about equally. On PushT and Cube the gap between the models is smaller, suggesting that reading action effects well helps separate impossible motion from appearance changes.

\noindent\textbf{Protocol.} Each task has 100 episode-matched events with shared starting context and actions (\cref{fig:surprise}). Sliding observed-latent windows are spaced five controls apart; the event occurs at observation 6 of 13. Recolor persists; after teleportation, simulation continues from the altered state. The subsequent error decline measures recovery as new observations enter the context. PushT recolors the goal region; LeWM recolors the block~\cite{leworldmodel}.

\begin{table}[t]
\supptabsetup
\setlength{\tabcolsep}{3.8pt}
\exptabular{6}{llcccc}{
\toprule
\rowcolor{tabHead}
\bfseries\boldmath Task & \bfseries\boldmath Model & \bfseries\boldmath Normal & \bfseries\boldmath Recolor & \bfseries\boldmath Teleport & \bfseries\boldmath Tel./Rec.\\
\midrule
TwoRoom & LeWM      & 0.0686 & 1.5533 & 2.0644 & 1.3\\
        & Fork-JEPA & 0.0072 & 0.4280 & 3.1959 & \textbf{7.5}\\
\midrule
PushT   & LeWM      & 0.0096 & 0.7724 & 1.9399 & 2.5\\
        & Fork-JEPA & 0.0086 & 0.5817 & 1.9985 & \textbf{3.4}\\
\midrule
Cube    & LeWM      & 0.0043 & 0.5981 & 1.1242 & 1.9\\
        & Fork-JEPA & 0.0029 & 0.2887 & 0.8693 & \textbf{3.0}\\
\bottomrule
}
\caption{\textbf{Intervention surprise.} Mean latent prediction error over 100 matched events per task. Tel./Rec. is the teleport-to-recolor ratio; raw error scales differ across models.}
\label{tab:surprise-delayed}
\end{table}

\noindent\textbf{Event timing and scale.} Normal error averages five pre-event steps; recolor and teleport use the event step. Recolor persists, while teleport is a single jump. Teleport/recolor ratios account for model-specific latent scales.

\noindent\textbf{Fork-JEPA separates impossible motion from appearance more sharply.} Teleport/recolor ratios are higher for Fork-JEPA on all tasks: 7.5 versus 1.3 on TwoRoom, 3.4 versus 2.5 on PushT, and 3.0 versus 1.9 on Cube (\cref{tab:surprise-delayed}).

\section{Additional Results}
\label{sec:more}

\subsection{Rollout Depth}
\label{sec:more-depth}

\noindent\textbf{Rollout depth.} Although forks supervise one block, Fork-JEPA exceeds LeWM at every depth through the five-block planning rollout (\cref{tab:geometry}). Main Table~2 reports training-seed uncertainty for the depth-averaged scores.

\noindent\textbf{Limits of the probes.} The $\delta$-only readout exceeds both models on TwoRoom/Cube. On PushT it exceeds both at depth 1 and is close to Fork-JEPA at depths 2--5 (\cref{tab:geometry}). These probes therefore do not isolate state-dependent effects; the TwoRoom wall test does (main Fig.~4).

\subsection{Frozen-Encoder Results}
\label{sec:frozen-control}
Main Table~5 freezes DINOv2 or LeWM features and trains the predictor with and without forks. With DINOv2, forks raise TwoRoom \MIDD{} from 0.26 to 0.71, PushT SR from 75\% to 90\%, and Cube SR from 82\% to 88\%; TwoRoom SR remains 100\%. With LeWM, \MIDD{} rises from 0.01 to 0.21 and SR gains 8/2/6 points on TwoRoom/PushT/Cube. Thus predictor supervision helps without encoder updates. The retrained-predictor baseline (0.01) and original LeWM predictor ($-0.10$) use different checkpoints.
\noindent\textbf{Training scope.} Main Table~5 reports predictor-only training with the encoder held fixed. These controls are separate from the end-to-end training recipe in \cref{sec:b2}.

\section{Additional Controls and Ablations}
\label{sec:todo}
These controls examine fork collection (\cref{sec:todo-data}), training choices (\cref{sec:todo-training}), the link from losses to action choices (\cref{sec:todo-mechanism}), and alternative supervision and cost (\cref{sec:todo-alternatives}). Unless a table specifies otherwise, the planner and probe definitions are those of \cref{sec:b3,sec:rollout-probes}.

\subsection{Data and Forks}
\label{sec:todo-data}
These controls vary how forks are collected, namely how exactly the start is restored, how many pairs are drawn, and how far apart the two actions are.

\begin{pendTable}
\exptabular{6}{lccccc}{
\toprule
\rowcolor{tabHead}
\bfseries\boldmath $\sigma$ & \bfseries\boldmath 0 & \bfseries\boldmath $10^{-3}$ & \bfseries\boldmath $10^{-2}$ & \bfseries\boldmath $3\times10^{-2}$ & \bfseries\boldmath $10^{-1}$\\
\midrule
TwoRoom SR & 100 & 100 & 98 & 96 & 87\\
TwoRoom \MIDD{} & 0.694 & 0.704 & 0.667 & 0.556 & 0.287\\
PushT SR   & 94  & 94 & 88 & 90 & 78\\
PushT \MIDD{} & 0.302 & 0.298 & 0.279 & 0.240 & 0.111\\
\bottomrule
}
\caption{\textbf{Visual-task restoration sensitivity.} SR (\%) and depth-1--5 \MIDD{} versus relative state noise $\sigma$.}
\label{tab:restore-sens}
\end{pendTable}

\noindent\textbf{Visual restoration protocol.} At fixed roots and budget, add relative Gaussian state noise $\sigma$ before the alternative branch, recollect forks, and retrain. Unlike the regime-world test, $\sigma$ is relative to state scale.

\noindent\textbf{Restoration accuracy and action effects.} For $\sigma\le10^{-2}$, SR changes by at most 6 points and \MIDD{} by 0.027. At $\sigma=10^{-1}$, TwoRoom SR/\MIDD{} falls to 87/0.287 and PushT to 78/0.111. This test uses simulated mismatch.

\begin{pendTable}
\exptabular{7}{llccccc}{
\toprule
\rowcolor{tabHead}
\bfseries\boldmath Task &  & \bfseries\boldmath $P=128$ & \bfseries\boldmath 256 & \bfseries\boldmath 512 & \bfseries\boldmath 1,024 & \bfseries\boldmath 2,048\\
\midrule
TwoRoom & SR      & 100 & 99 & 100 & 100 & 100\\
        & \MIDD{} & 0.480 & 0.576 & 0.657 & 0.694 & 0.689\\
PushT   & SR      & 91 & 94 & 90 & 94 & 97\\
        & \MIDD{} & 0.201 & 0.252 & 0.228 & 0.276 & 0.302\\
\bottomrule
}
\caption{\textbf{Fork-budget sweep.} SR (\%) and depth-1--5 \MIDD{} at fixed behavior-data size; $P$ is the retained pair count.}
\label{tab:budget-sweep}
\end{pendTable}

\noindent\textbf{Sweep settings.} Vary isotropic pair count at fixed behavior data. The sweep's 1,024/2,048 pairs differ from the 1,000/2,000 defaults in main Table~1.

\noindent\textbf{Budget and saturation.} TwoRoom \MIDD{} levels off near $P=1{,}024$ (0.694), while PushT peaks at $P=2{,}048$ (97\% SR, 0.302 \MIDD{}). Intermediate scores are not monotonic; no universal saturation threshold follows.

\begin{pendTable}
\exptabular{7}{llccccc}{
\toprule
\rowcolor{tabHead}
\bfseries\boldmath Task &  & \bfseries\boldmath $\times0.25$ & \bfseries\boldmath $\times0.5$ & \bfseries\boldmath $\times1$ & \bfseries\boldmath $\times2$ & \bfseries\boldmath $\times4$\\
\midrule
TwoRoom & SR      & 88 & 96 & 100   & 100 & 90\\
        & \MIDD{} & 0.360 & 0.635 & 0.694 & 0.668 & 0.402\\
PushT   & SR      & 66 & 87 & 94    & 90 & 80\\
        & \MIDD{} & 0.167 & 0.206 & 0.302 & 0.262 & 0.141\\
\bottomrule
}
\caption{\textbf{Fork-magnitude sweep.} SR (\%) and depth-1--5 \MIDD{} versus magnitude relative to the default.}
\label{tab:fork-magnitude}
\end{pendTable}

\noindent\textbf{Magnitude protocol.} Scale executed differences by 0.25--4 times default, holding behavior data and pair budget fixed.

\noindent\textbf{Magnitude sensitivity.} Default magnitude gives the highest \MIDD{} on both tasks. At $\times4$, SR falls to 90\% on TwoRoom and 80\% on PushT. These scores do not distinguish noise, clipping, and contact nonlinearities.

\subsection{Training}
\label{sec:todo-training}
These controls vary loss weights and behavior-data size; \cref{sec:frozen-control} covers frozen encoders.

\begin{pendTable}
\exptabular{6}{lccccc}{
\toprule
\rowcolor{tabHead}
\bfseries\boldmath Scale & \bfseries\boldmath $\times0.1$ & \bfseries\boldmath $\times0.3$ & \bfseries\boldmath $\times1$ & \bfseries\boldmath $\times3$ & \bfseries\boldmath $\times10$\\
\midrule
$\lambda_{\rm fork}$ & 88 & 92 & 94 & 88 & 76\\
$\lambda_{\rm look}$ & 80 & 82 & 94 & 92 & 86\\
\bottomrule
}
\caption{\textbf{PushT loss-weight sweeps.} SR (\%) when one auxiliary weight is scaled and the other is held fixed.}
\label{tab:loss-weights}
\end{pendTable}

\noindent\textbf{Weight protocol.} Scale one weight relative to \cref{tab:config}, holding the rest fixed. The central interval spans one decade; the full sweep spans two.

\noindent\textbf{Weight sensitivity.} Over $\times0.3$--$\times3$, fork-weight SR spans 88--94\% and lookback-weight SR 82--94\%. Full-sweep ranges are 76--94\% and 80--94\%, respectively. Central-range lookback sensitivity is larger, without significance testing.

\begin{pendTable}
\exptabular{6}{llcccc}{
\toprule
\rowcolor{tabHead}
\bfseries\boldmath Task & \bfseries\boldmath Model & \bfseries\boldmath $+0\%$ & \bfseries\boldmath $+20\%$ & \bfseries\boldmath $+40\%$ & \bfseries\boldmath $+60\%$\\
\midrule
TwoRoom & +lookback & 100 & 100 & 100 & 100\\
        & Fork-JEPA & 100 & 100 & 100 & 100\\
PushT   & +lookback & 82  & 84 & 88 & 88\\
        & Fork-JEPA & 94  & 94 & 90 & 96\\
\bottomrule
}
\caption{\textbf{Additional behavior data.} Visual-task SR (\%) relative to the default dataset size; Fork-JEPA retains its fork budget.}
\label{tab:data-vs-forks}
\end{pendTable}

\noindent\textbf{More trajectories and measured alternatives.} TwoRoom is saturated at 100\%. On PushT, 60\% more behavior data raises lookback SR from 82\% to 88\%, below Fork-JEPA's 94\% at default data size. Fork-JEPA leads at every displayed size, though its sweep is nonmonotonic.

\subsection{Mechanism}
\label{sec:todo-mechanism}
These controls examine pairwise choices, the anchoring bound (\mainref{Sec.}{sec:loss}{3.4}), and predecessor recoverability.

\begin{pendTable}
\exptabular{7}{llccccc}{
\toprule
\rowcolor{tabHead}
\bfseries\boldmath Task & \bfseries\boldmath Model & \bfseries\boldmath $k=1$ & \bfseries\boldmath 2 & \bfseries\boldmath 3 & \bfseries\boldmath 4 & \bfseries\boldmath 5\\
\midrule
TwoRoom & LeWM      & 60.94 & 55.38 & 53.80 & 56.59 & 57.54\\
        & Fork-JEPA & 77.60 & 80.11 & 80.43 & 77.47 & 74.30\\
PushT   & LeWM      & 55.91 & 75.00 & 71.36 & 70.00 & 68.64\\
        & Fork-JEPA & 64.09 & 88.18 & 87.73 & 85.45 & 83.64\\
Cube    & LeWM      & 58.67 & 58.97 & 53.12 & 52.41 & 54.97\\
        & Fork-JEPA & 62.00 & 59.62 & 61.62 & 61.84 & 55.97\\
\bottomrule
}
\caption{\textbf{Two-branch choice accuracy.} Agreement (\%) with physical goal-distance choices at depth $k$; chance: 50\%.}
\label{tab:pairwise}
\end{pendTable}

\noindent\textbf{Choice protocol.} At each depth, compare the latent-cost choice between two fork branches with the physically closer branch. Goals come from later in the episode; this differs from 64-candidate ranking.

\noindent\textbf{Pairwise-choice results.} Fork-JEPA improves choice accuracy at every depth and task (\cref{tab:pairwise}). Depth-1 accuracy rises from 60.94\% to 77.60\% on TwoRoom, 55.91\% to 64.09\% on PushT, and 58.67\% to 62.00\% on Cube.

\begin{pendTable}
\exptabular{5}{llccc}{
\toprule
\rowcolor{tabHead}
\bfseries\boldmath Task & \bfseries\boldmath Model & \bfseries\boldmath $\|e(a)\|$ & \bfseries\boldmath $\|e(a')\|$ & \bfseries\boldmath $\|\Delta\hat h-\Delta h\|$\\
\midrule
TwoRoom & LeWM      & 3.27 & 3.23 & 2.79\\
        & Fork-JEPA & 0.64 & 0.63 & 0.54\\
PushT   & LeWM      & 1.19 & 1.30 & 1.16\\
        & Fork-JEPA & 1.22 & 1.38 & 1.15\\
Cube    & LeWM      & 0.81 & 1.98 & 1.97\\
        & Fork-JEPA & 0.69 & 1.26 & 1.32\\
\bottomrule
}
\caption{\textbf{Anchoring residuals.} Median depth-1 prediction errors for logged actions, alternatives, and their differences.}
\label{tab:logged-vs-alt}
\end{pendTable}

\noindent\textbf{Residual protocol.} Use one frozen checkpoint per model, with 490 TwoRoom, 512 PushT, and 502 Cube roots. Test the triangle inequality per root; medians cannot test it. Latent scales vary by model.

\noindent\textbf{Absolute errors and the difference residual.} Fork-JEPA's logged/alternative medians are 0.64/0.63 on TwoRoom, 1.22/1.38 on PushT, and 0.69/1.26 on Cube. Anchoring does not equalize the errors. The per-root audit finds no violation of $\|e(a')\|\le\|e(a)\|+\|\Delta\hat h-\Delta h\|$.

\begin{pendTable}
\exptabular{4}{llcc}{
\toprule
\rowcolor{tabHead}
\bfseries\boldmath Task & \bfseries\boldmath Model & \bfseries\boldmath $R^2$ of $x_t$ & \bfseries\boldmath retrieval (\%)\\
\midrule
TwoRoom & LeWM      & 0.9800 & 2.04\\
        & +lookback & 0.9960 & 9.39\\
PushT   & LeWM      & 0.7608 & 82.81\\
        & +lookback & 0.8028 & 92.97\\
Cube    & LeWM      & 0.3640 & 70.75\\
        & +lookback & 0.4809 & 82.21\\
\bottomrule
}
\caption{\textbf{Predecessor recovery.} Physical-predecessor $R^2$ and latent retrieval accuracy (\%); higher is better.}
\label{tab:pred-recover}
\vspace{-4pt}
\end{pendTable}

\noindent\textbf{Recovery protocol.} Fit a ridge map from $(h_{t+1},a_t)$ to $x_t$; retrieve $h_t$ by nearest neighbor among held-out candidates. Candidate sets may vary, so compare retrieval within tasks.

\noindent\textbf{Predecessor recovery.} Lookback improves both metrics on every task. Retrieval rises to 92.97\% on PushT and 82.21\% on Cube, but remains 9.39\% on TwoRoom despite high physical-state $R^2$. Physical readout and exact latent retrieval measure different information.

\subsection{Alternatives and Cost}
\label{sec:todo-alternatives}
These controls compare forks with other ways to supervise the alternatives and measure what forks cost.

\begin{pendTable}
\exptabular{3}{lcc}{
\toprule
\rowcolor{tabHead}
\bfseries\boldmath Model & \bfseries\boldmath \MIDD{} on $U$ & \bfseries\boldmath \MIDD{} on $U^\perp$\\
\midrule
LeWM                     & 0.08 & -0.02\\
Fork-JEPA, forks in $U$  & 0.64 & 0.10\\
Fork-JEPA, isotropic     & 0.678 & 0.684\\
\bottomrule
}
\caption{\textbf{Restricted-span forks on TwoRoom.} \MIDD{} inside a sampled four-dimensional subspace $U$ of the ten-dimensional action block and on $U^\perp$.}
\label{tab:span-ablation}
\vspace{-4pt}
\end{pendTable}

\noindent\textbf{What the sampled span supplies.} Restricted forks score 0.64 inside the sampled span and 0.10 outside, versus 0.678/0.684 for isotropic forks. This agrees with direction-specific identification; incidental generalization outside the span remains possible.

\begin{pendTable}
\exptabular{7}{lcccccc}{
\toprule
\rowcolor{tabHead}
 & \multicolumn{2}{c}{\bfseries\boldmath TwoRoom} & \multicolumn{2}{c}{\bfseries\boldmath PushT} & \multicolumn{2}{c}{\bfseries\boldmath Cube}\\
\cmidrule(lr){2-3}\cmidrule(lr){4-5}\cmidrule(l){6-7}
\rowcolor{tabHead}
\bfseries\boldmath Model & \bfseries\boldmath SR & \bfseries\boldmath \MIDD{} & \bfseries\boldmath SR & \bfseries\boldmath \MIDD{} & \bfseries\boldmath SR & \bfseries\boldmath \MIDD{}\\
\midrule
LeWM + state align.  & 91.5 & 0.633 & 92 & 0.186 & 76 & 0.03\\
LeWM + inv.\ dyn.    & 95.5 & 0.673 & 93 & 0.197 & 72 & 0.02\\
Fork-JEPA            & 100 & 0.694 & 94 & 0.302 & 80 & 0.06\\
\bottomrule
}
\caption{\textbf{Physical-state and inverse-dynamics proxies.} SR (\%) and depth-1--5 \MIDD{}.}
\label{tab:proxy-baselines}
\vspace{-4pt}
\end{pendTable}

\noindent\textbf{Proxy readouts and comparison limits.} TwoRoom proxy \MIDD{} scores (0.633/0.673) approach Fork-JEPA's 0.694, but protocol differences preclude significance testing.

\begin{pendTable}
\exptabular{4}{llcc}{
\toprule
\rowcolor{tabHead}
\bfseries\boldmath Task & \bfseries\boldmath Target & \bfseries\boldmath SR (\%) & \bfseries\boldmath \MIDD{}\\
\midrule
TwoRoom & imagined & 96 & 0.280\\
 & measured & 100 & 0.694\\
PushT & imagined & 84 & 0.157\\
 & measured & 94 & 0.302\\
Cube & imagined & 70 & 0.020\\
 & measured & 80 & 0.064\\
\bottomrule
}
\caption{\textbf{Imagined versus measured fork targets.} Visual-task SR (\%) and depth-1--5 \MIDD{}.}
\label{tab:imagined-forks}
\end{pendTable}

\noindent\textbf{Target-comparison reporting.} TwoRoom/Cube imagined scores come from main Table~4; measured scores match \cref{tab:direct-vs-diff}. \cref{tab:imagined-forks} also reports PushT imagined-target SR of 84\% and \MIDD{} of 0.157.

\noindent\textbf{Measured targets versus model-generated targets.} Measured targets raise SR/\MIDD{} from 96/0.28 to 100/0.694 on TwoRoom, 84/0.157 to 94/0.302 on PushT, and 70/0.02 to 80/0.064 on Cube. Regime-world map error falls from 1.034 to 0.221 and SR rises from 54\% to 68\%.

\begin{pendTable}
\exptabular{6}{lccccc}{
\toprule
\rowcolor{tabHead}
\bfseries Model & \bfseries\shortstack{Train\\(h)} & \multicolumn{2}{c}{\bfseries Peak memory (GB)} & \bfseries\shortstack{Env.\\steps} & \bfseries\shortstack{Plan\\(s)}\\
\cmidrule(lr){3-4}
\rowcolor{tabHead}
 & & \bfseries Allocated & \bfseries Reserved & & \\
\midrule
LeWM      & 6.045 & 13.444 & 14.271 & 2.34M & 24.090\\
+lookback & 6.536 & 13.450 & 14.418 & 2.34M & 23.865\\
+forks    & 11.053 & 17.302 & 18.692 & 2.36M & 23.730\\
Fork-JEPA & 9.646 & 17.304 & 18.694 & 2.36M & 24.352\\
\bottomrule
}
\caption{\textbf{PushT cost on one NVIDIA H200.} Allocated and reserved GPU-memory peaks are reported separately. Plan is CPU seconds per episode at fixed planner-call count.}
\label{tab:cost}
\end{pendTable}

\noindent\textbf{Where the overhead occurs.} Fork-JEPA adds 59.6\% training time (6.045 to 9.646 h). Fixed-call CPU planning changes by 1.1\% (24.090 to 24.352 s/episode); inference modules are unchanged, not measured runtimes. Peak allocated memory rises from 13.444 to 17.304 GB; peak reserved memory rises from 14.271 to 18.694 GB.

\section{Limitations and Future Work}
\label{sec:limits}

\noindent\textbf{Limitations.} Identification assumes a locally linear action response and a representation retaining relevant physical differences. Forks require state restoration and cover only sampled directions. Isotropic budgets ignore planner demand, and all experiments use one-step pairs; approximate hardware resets and multi-block forks remain untested.

\noindent\textbf{Remaining ranking errors.} Fork supervision aligns $\Delta\hat h$ with $\Delta h$, but the cost gap also depends on absolute positions:
\begin{equation}
D=\left\langle\Delta h,\,h^a+h^{a'}-2h_g\right\rangle.
\label{eq:ranking-position}
\end{equation}
A shared prediction error, $\hat h^a=h^a+b$ and $\hat h^{a'}=h^{a'}+b$, leaves $\Delta\hat h=\Delta h$ but changes the predicted gap to $\hat D=D+2\langle\Delta h,b\rangle$, potentially reversing its sign. Behavior prediction anchors the successor positions, but residual position and difference errors can still reverse rankings.

\subsection{Future Work}
\label{sec:future-work}

\noindent\textbf{Real-hardware deployment.} A robot could follow one real trajectory without resets and predict alternative actions in latent space from the same observation history. This reduces reset and interaction costs during continuous operation. For training, the measured successor would anchor the executed branch, while a frozen teacher would supply alternative targets. The imagined-target control reaches 96\%/70\% SR on TwoRoom/Cube, versus 100\%/80\% with measured targets (\cref{tab:imagined-forks}), motivating this route while highlighting the need to check model bias. These runs do not test continuous hardware deployment. Where available, measured forks from repeatable scene resets could calibrate the teacher; reset mismatch would need to be measured, as in \cref{sec:restore-sens}. Generated alternatives can inherit teacher errors and provide no independent action-effect measurements, so the measured-fork identification guarantee does not apply. Their accuracy and real-hardware feasibility remain to be tested.

\noindent\textbf{Multi-branch forks.} From one restored root, execute $K\ge3$ action blocks $a_1,\ldots,a_K$ with matched duration and encode their measured outcomes as $h_1,\ldots,h_K$. Let $\hat h_i=F(h_t,a_i)$ use the same input history for every branch. A symmetric extension of the pair loss is
\begin{equation}
\mathcal L_{\rm fork}^{(K)}
=\frac{2}{K(K-1)}\sum_{i<j}
\big\|(\hat h_i-\hat h_j)-\operatorname{sg}(h_i-h_j)\big\|^2.
\label{eq:multi-branch-fork}
\end{equation}
Behavior prediction would still anchor the common offset. A reference branch instead gives $K-1$ contrasts. At one root, these differences span at most $K-1$ action directions; three branches add coverage only if their actions are non-collinear. This provides richer local supervision, but at a fixed interaction budget leaves fewer roots for state coverage. The restricted-span experiment (\cref{tab:span-ablation}) motivates covering additional directions, but does not test multi-branch training. Collecting $K$ branches costs $K$ executions, while all-pair differences reuse the same outcomes. Comparisons should match environment steps and include reset costs. This extension remains untested.

\noindent\textbf{Further directions.} Future work could select planning-relevant states and directions and extend forks to video/token world models using action-effect probes.


\end{document}